\documentclass[lettersize,journal]{IEEEtran}
\usepackage{amsmath,amsfonts}
\usepackage{algorithmic}
\usepackage{algorithm}
\usepackage{array}
\usepackage{textcomp}
\usepackage{stfloats}
\usepackage{url}
\usepackage{verbatim}
\usepackage{graphicx}
\usepackage{cite}
\usepackage{amsmath}
\usepackage{amsthm}
\usepackage{booktabs}
\usepackage{algorithm}
\usepackage{algorithmic}
\usepackage[switch]{lineno}
\usepackage{color}
\usepackage{colortbl}
\usepackage{amssymb}
\usepackage{subfigure}
\usepackage{enumerate}
\usepackage{microtype}
\usepackage{multirow}
\usepackage{bm}
\usepackage{booktabs} % 用于提供更好的表格线条
\begin{document}

\title{Towards Zero-shot Human-Object Interaction Detection via Vision-Language Integration}

\author{Weiying Xue, Qi Liu*,~\IEEEmembership{Senior Member,~IEEE,} Qiwei Xiong, Yuxiao Wang, Zhenao Wei, \\ Xiaofen Xing, Xiangmin Xu
        % <-this % stops a space
% \thanks{This paper was produced by the IEEE Publication Technology Group. They are in Piscataway, NJ.}% <-this % stops a space
% \thanks{Manuscript received April 19, 2021; revised August 16, 2021.}
\thanks{* Corresponding author: Qi Liu (drliuqi@scut.edu.cn)}
\thanks{Weiying Xue (202320163283@mail.scut.edu.cn), Qi Liu, Qiwei Xiong (xiongqiwei@vip.qq.com), Yuxiao Wang (ftwangyuxiao@mail.scut.edu.cn), Zhenao Wei (wza@scut.edu.cn), Xiaofen Xing (xfxing@scut.edu.cn), and Xiangmin Xu (xmxu@scut.edu.cn) are with the School of Future Technology, South China University of Technology, China 511400.}
}

% The paper headers
\markboth{Journal of \LaTeX\ Class Files,~Vol.~14, No.~8, August~2021}%
{Shell \MakeLowercase{\textit{et al.}}: A Sample Article Using IEEEtran.cls for IEEE Journals}

\IEEEpubid{0000--0000/00\$00.00~\copyright~2021 IEEE}
% Remember, if you use this you must call \IEEEpubidadjcol in the second
% column for its text to clear the IEEEpubid mark.

\maketitle

\begin{abstract}
Human-object interaction (HOI) detection aims to locate human-object pairs and identify their interaction categories in images. Most existing methods primarily focus on supervised learning, which relies on extensive manual HOI annotations. In this paper, we propose a novel framework, termed Knowledge Integration to HOI (KI2HOI), that effectively integrates the knowledge of visual-language model to improve zero-shot HOI detection. Specifically, the verb feature learning module is designed based on visual semantics, by employing the verb extraction decoder to convert corresponding verb queries into interaction-specific category representations. We develop an effective additive self-attention mechanism to generate more comprehensive visual representations. Moreover, the innovative interaction representation decoder effectively extracts informative regions by integrating spatial and visual feature information through a cross-attention mechanism. To deal with zero-shot learning in low-data, we leverage \textit{a priori} knowledge from the CLIP text encoder to initialize the linear classifier for enhanced interaction understanding. Extensive experiments conducted on the mainstream HICO-DET and V-COCO datasets demonstrate that our model outperforms the previous methods in various zero-shot and full-supervised settings.
\end{abstract}

\begin{IEEEkeywords}
Human-object interaction, Knowledge Integration, Zero-shot, Weakly Supervision.
\end{IEEEkeywords}

\section{INTRODUCTION}
\IEEEPARstart{H}{uman-object} interaction (HOI) detection is a process of detecting interaction between a human and an object in an image~\cite{antoun2022human}. Precisely estimating human-object interactions can greatly improve various visual understanding tasks, such as image retrieval~\cite{johnson2015image}, visual question answering~\cite{chen2020counterfactual}, and scene graph generation~\cite{R3D,9996418}. Given a series of $\langle$``Human", ``Object", ``Verb"$\rangle$ triples, an HOI detector is needed to locate human-object pairs and identify their interactions. However, most HOI detectors typically require a significant number of pre-defined HOI categories. Considering the diversity and complexity of human-object interaction in the real world, it is time-consuming and laborious to define all-natural interaction annotations in advance manually.
 \begin{figure}[!ht]
    \centering
    \includegraphics[width=\linewidth]{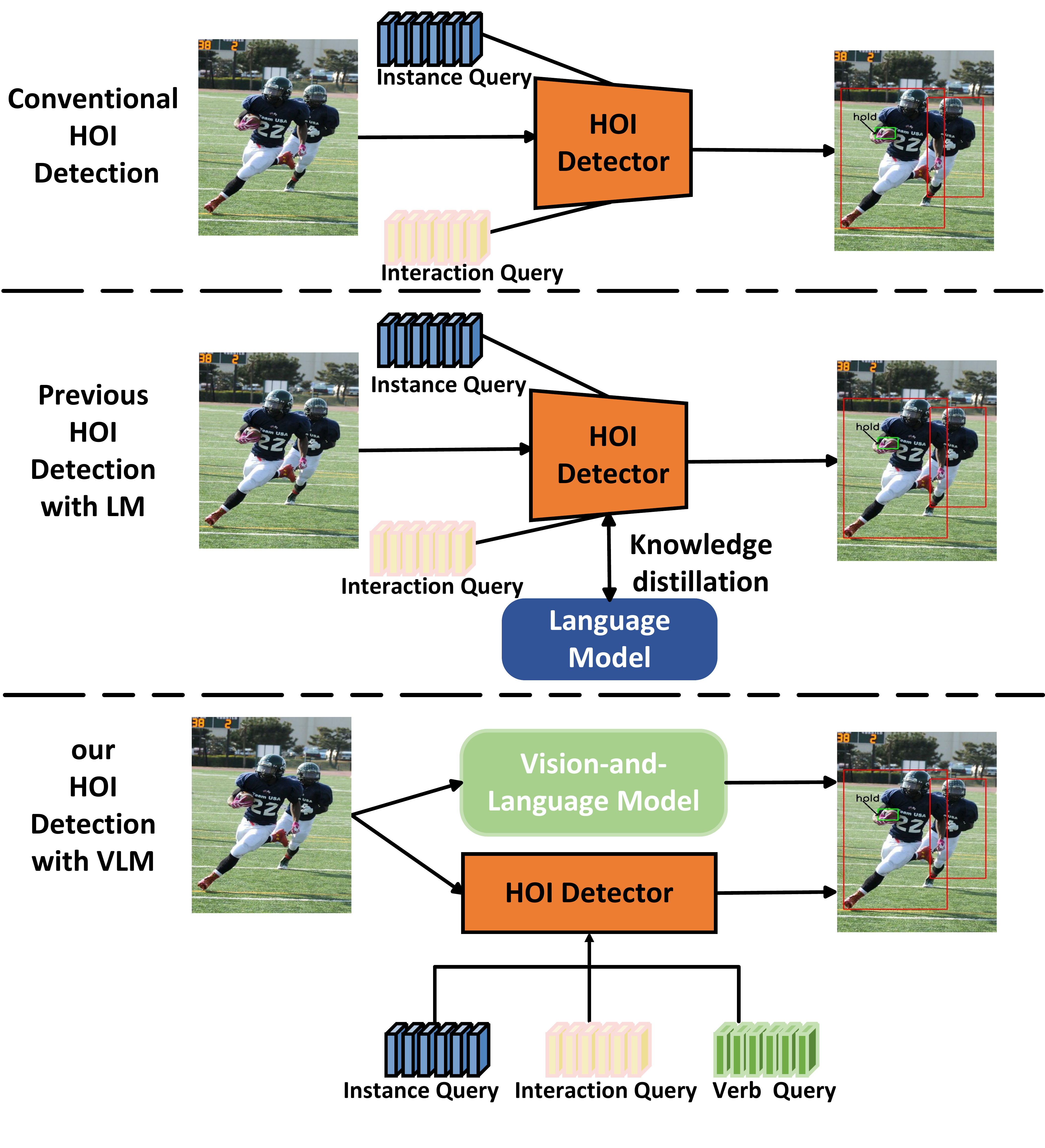}
    \caption{Comparison of HOI detection. Conventional HOI detection required manually annotated datasets for training. Previous HOI detection with language model (LM) employed limited knowledge distillation to visual detectors, but it is limited to handling potential interactions among unseen human-object pairs. Our model fully leverages visual-language model (VLM) and verb queries for effective knowledge integration, to promote unseen interaction recognition.}
    \label{fig1}
\end{figure} 

The development of pre-trained visual-language models on large-scale data has significantly accelerated the progress in studying zero-shot learning~\cite{du2022learning,DBLP:journals/corr/abs-2305-07011}, especially CLIP~\cite{radford2021learning} demonstrates remarkable transfer ability in various subsequent tasks. The most recent zero-shot HOI detectors leverage the comprehensive visual and linguistic knowledge of CLIP to detect novel HOIs. For example, the knowledge of pre-trained visual language model is transferred to EOID~\cite{wu2023end} and DOQ~\cite{qu2022distillation} via knowledge distillation to achieve zero-shot HOI detection. Previous approaches utilize cross-modal knowledge in a too-restrictive manner, failing to fully capitalize on the potential of cross-modal information and Language Models in the field of Human-Object Interaction (HOI) detection. Knowledge distillation depends on the quality of the teacher model and the used training data. When the training process does not include unknown categories, the distillation process may be biased towards known category samples, thereby the generalization ability is limited.

\IEEEpubidadjcol 
Considering the recent outstanding performance of visual-language models in various fundamental tasks, as depicted in Figure~\ref{fig1}, we propose a novel one-stage zero-shot framework for Human-Object Interaction detection, named Knowledge Integration to HOI (KI2HOI). We harness the capabilities of foundational models to enhance the comprehension of complex interactive semantics within visual data. Instead of using knowledge distillation on CLIP for detectors, KI2HOI learns \textit{a priori} knowledge from CLIP and integrates compositional features extracted from multiple perspectives in HOI detection. On the one hand, leveraging the off-the-shelf object detector, Ho-Pair Encoder is designed to extract rich local and global visual features. On the other hand, motivated by the Query2label~\cite{liu2021query2label}, a transformer-based model utilized for multi-label image classification, we extend to extract verb representations based on visual semantics for HOI classification in conjunction with CLIP, where each query is associated with a specific interaction category in training and inference. Such queries adaptively focus on relevant interactive regions to understand and represent each interaction, yielding better intricate. Furthermore, to address the long-tail distribution problem, we combine the verb features with interaction semantic representations and our interaction representation decoder consists of a verb representation head and a set of class weights computed via text semantics algorithm. While traditional interaction classification heads work well in handling visible interactions, they fall short in tackling invisible ones. Thus, we combine the prompt-based verb classification head with the redesigned verb representation head to enhance the HOI prediction.

To summarize, our contributions are as follows:
\begin{itemize}
\item We propose a novel framework, named KI2HOI, for zero-shot HOI detection that directly retrieves visual and linguistic knowledge of CLIP. Our KI2HOI effectively utilizes \textit{a priori} knowledge and achieves superior zero-shot transferability.
\item We develop a verb feature extraction strategy to explicitly match verb queries and interactions for more expressive intricate representations between verbs and their associated interactions.
\item We conduct extensive experiments on HICO-Det for the zero-shot learning task and perform additional comprehensive experiments on HICO-Det and V-COCO for the supervised learning task. Our model outperforms the state-of-the-art methods in zero-shot and full-supervised settings, establishing a new state-of-the-art.
\end{itemize}

\section{RELATED WORKS}
\subsection{Human-Object Interaction Detection}
HOI detection methods can be roughly categorized as two-stage or one-stage solutions. Two-stage methods~\cite{wan2023weakly,li2022improving,liu2020amplifying,cao2023re,9927451,10415065} first detect all candidate interaction pairs and then feed to CNN network to predict the interaction relationships between candidate human-object pairs. Most two-stage detection models use existing object detection models and prioritize improving interaction prediction models. FCL~\cite{hou2021detecting} proposed an object fabricator to generate effective object representations, which were then combined with verbs to compose new HOI samples, thus increasing the diversity of training data. UPT~\cite{zhang2022efficient} applied a unary-pairwise transformer to represent each target's instance details as unary and pairwise representations. In comparison to two-stage methods, one-stage solutions~\cite{9878997,kim2021hotr,liu2022interactiveness,wu2023end,yuan2022rlip,kim2023relational,zhou2022human,9570360} captured context information during the early stage of feature extraction, leading to improved HOI detection performance. The success of DETR~\cite{carion2020end} has inspired many researchers in studying HOI detection QPIC~\cite{tamura2021qpic} applied additional detection heads and relied on a bipartite graph matching algorithm to locate HOI instances and identify interactions. EOID~\cite{wu2023end} developed a teacher-student model and designed a two-stage Hungarian matching algorithm. RR-Net~\cite{9570360} introduced a relation-aware frame to build progressive structure for interaction inference, which
imitates the human visual mechanism of recognizing HOI by comprehending visual instances and interactions coherently. HOICLIP~\cite{10204103} proposed a new transfer strategy that used visual semantic algorithms to represent verbs. Our work belongs to a one-stage end-to-end approach to study HOI detection.
\subsection{Vision-and-Language Pre-training}
The advanced vision-and-language pre-training (VLP) multimodal learning framework can acquire generalized multimodal representations from large-scale image and text data~\cite{li2022grounded}. It has a wide application in the fields of multimodal retrieval~\cite{dzabraev2021mdmmt,10382576}, visual and language navigation~\cite{anderson2018vision}, image description~\cite{jin2021good}, and so on. Through effective cross-modal semantic alignment, particularly fine-grained semantic alignment, VLP contributes to cross-modal learning and generalization. Visual-language models succeed in enabling zero-shot open-vocabulary tasks from natural language supervision~\cite{gu2021open}, which inspires us to apply visual-language models for zero-shot HOI detection tasks.
\subsection{Zero-shot HOI}
Zero-shot HOI detection aims to generalize to unseen HOI categories during training effectively. Since the majority of HOI exhibit a long-tail distribution, attributed to the compositional nature of HOIs, prior research~\cite{9010418,hou2021detecting,liu2020consnet,hou2021affordance,radford2021learning,yuan2022rlip} on zero-shot HOI detection focuses on transferring knowledge from known HOI concepts to unseen classes. They can be categorized into three scenarios: unseen object, unseen action, and unseen combination. There exist primarily two research streams for addressing this task.
One stream~\cite{hou2021detecting,liu2020consnet,hou2021affordance} employed a combined learning approach for zero-shot HOI detection, which entailed separating HOI representations and combining known features to identify unseen HOI concepts. ConsNet~\cite{liu2020consnet}, for instance, constructed a consistency graph with both visual features of potential human-object pairs and word embeddings of HOI labels. With the advancement of multimodal learning, there is a growing interest in transferring knowledge from pre-trained visual language models, \textit{e.g.} CLIP is used to extract text embeddings of HOI descriptions for HOI detection tasks~\cite{radford2021learning}. RLIP~\cite{yuan2022rlip} proposed a transferable HOI detector via natural language supervision. Building upon GEN-VLKT~\cite{9878997}, HOICLIP~\cite{10204103} mapped image and text encodings to a joint visual-semantic space, to capture their correlations and effectively transfer knowledge from CLIP. Our work seeks to explore a more effective framework to make full use of CLIP for improving zero-shot HOI detection performance.

\section{METHODS}
In this section, we introduce the details of the proposed framework to improve its generalization ability in zero-shot HOI detection. In Section \ref{sec-3-1}, the overall architecture is presented. In Section \ref{sec-3-2}, we introduce our visual encoder and a strategy for enhancing global feature extraction. Section \ref{sec-3-3} presents verb feature learning based on verb queries. In section \ref{sec-3-4}, we design an interaction semantic representation for transferring knowledge to HOI detection. Section \ref{sec-3-5} describes the training and inference procedures of our model.

\begin{figure*}[!t]
    \centering
    \includegraphics[width=1\textwidth]{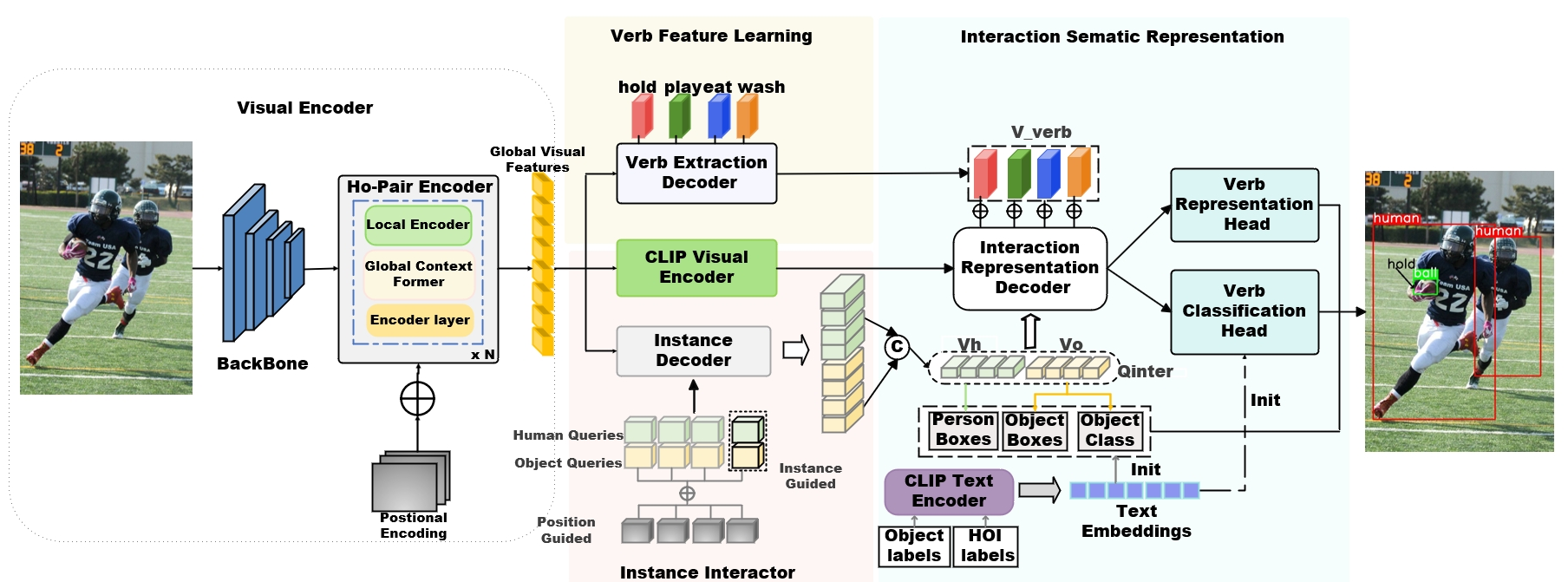}
    \caption{Overview of KI2HOI pipeline. It consists of four parts: visual encoder, verb feature learning, instance interactor, and interaction semantic representation (ISR). Given an image, firstly, we obtain the feature map through the backbone and then use our dedicated visual encoder to extract contextual global features. The instance interactor injects CLIP spatial information and global features to locate human-object pairs and classify object categories. In the verb feature learning module, associated verb queries are fed to the verb extraction decoder to obtain fine-grained verb features. The interaction semantic representation model inputs the verb features and the interaction features from encoders to extract the interaction representation.}
    \label{fig2}
\end{figure*} 
\subsection{Overall Architecture}\label{sec-3-1}
As shown in Figure~\ref{fig2}, our model consists of four primary components: visual encoder, verb feature learning, instance interactor, and interaction semantic representation. Given an input image \bm{$I$}, we initially extract feature maps via the backbone DETR~\cite{carion2020end}. The feature maps are subsequently input into the HO-pair Encoder to generate global visual feature $\bm{V}_{g}$, similar to GEN-VLKT~\cite{9878997}. Human query $\bm{Q}_{h}$ and object query $\bm{Q}_{o}$ are inputted into the instance interactor to compute the mean for both types of queries in the corresponding decoder layer. These outputs queries in the last decoder are then fed to classifiers which initialize by label’s text weights from CLIP to predict the interacting human bounding box $\bm{B}_{h} {\in} \mathbb{R}^{N \times4}$ and object bounding box $\bm{B}_{o} {\in} \mathbb{R}^{N \times4}$, where $N$ is the number of queries, and object class $\bm{{C}_{o}}{\in} \mathbb{R}^{N\times{C}}$, where \bm{$C$} denotes the object category. 

Furthermore, possible types of interactions depend on action classes. For example, humans are more likely to catch or play sports ball than to bite a sports ball. To reflect this characteristic of HOIs, our verb feature learning module uses a novel verb extraction decoder to perform verb recognition, as auxiliary information. We extract the spatial feature $\bm{V}_{sp}$ from the pre-trained CLIP Visual Encoder as memory, feed into the interaction representation decoder by the cross-attention mechanism to augment the interaction representation and recognition. Finally, the HOI prediction categories are generated by the output of a linear classifier. The details of each component are explained in the following sections.
 \begin{figure}[t]
    \centering
    \includegraphics[width=1\linewidth]{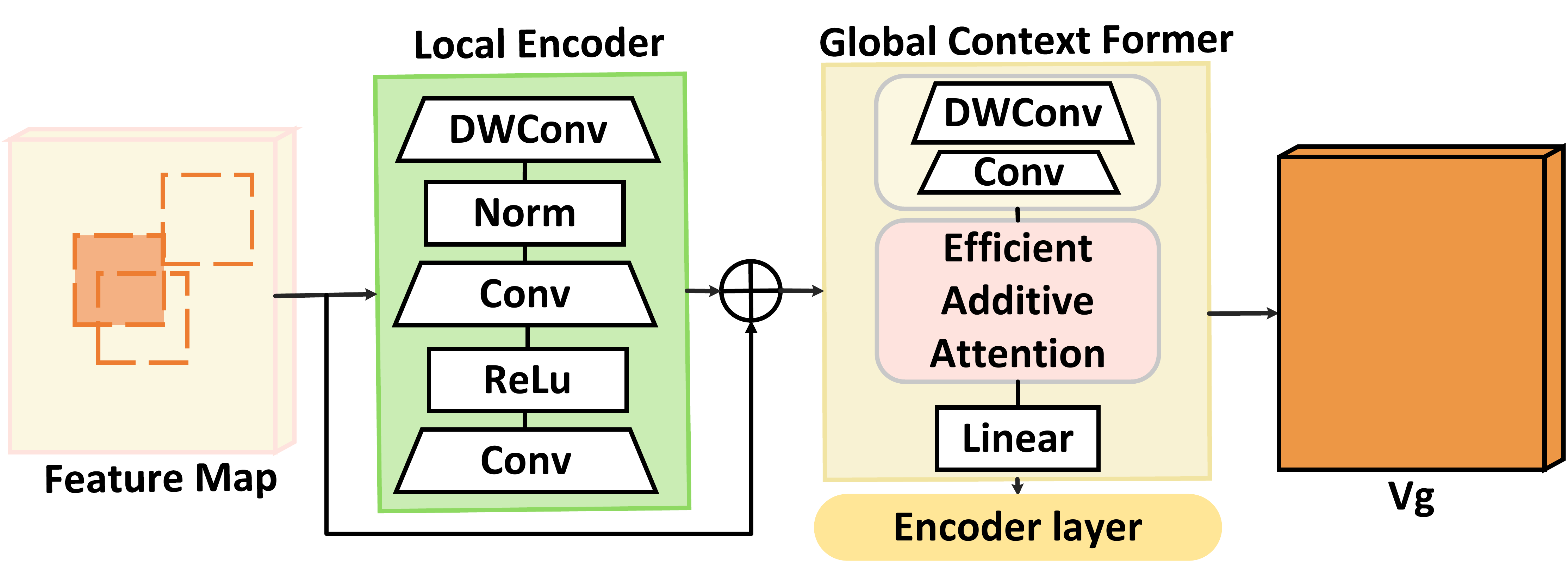}
    \caption{Structure of Ho-Pair Encoder. The local encoder is specifically engineered to encode efficient local characteristics, followed by $3\times3$ depth-wise convolution and two $ 1\times1$ convolutions for channel blending. The global context former is intended to capture comprehensive local-global representations by extracting local features from the local convolutional layers, an efficient additive attention module, and linear layers.}
    \label{fig3}
\end{figure} 
\subsection{Visual Encoder}\label{sec-3-2}
Our Ho-Pair Encoder is designed to significantly enhance performance by effectively addressing two critical design constraints inherent in EfficientFormer~\cite{li2022efficientformer}. The original EfficientFormer leverages 4D MetaBlocks, inspired by PoolFormer, for efficient learning of local image representations and 3D MetaBlocks, which utilize self-attention for capturing global context.  However, its performance is compromised by suboptimal token mixing and the restricted use of 3D MetaBlocks in the final stage due to the quadratic complexity of the multi-head self-attention (MHSA) mechanism, resulting in a potentially inconsistent and inadequate representation of context. To surmount these limitations, the Ho-Pair Encoder incorporates a Global Context Former that significantly enhances token mixing. This modification is designed to enhance the model's ability to integrate and process visual information across different scales and contexts, thereby providing a more robust and comprehensive understanding of the input data. We resize the cropped feature map obtained from backbone to a $7\times7$ grid using ROI-Align and then feed them into the Ho-Pair Encoder to acquire comprehensive high-level contextual information, as depicted in Figure~\ref{fig3}. The Ho-Pair Encoder comprises a local encoder and a global context former. Specifically, the feature map $\bm{V}_{F}$ is input to a $3\times3$ depth-wise convolution (DWConv) with batch normalization (BN). The results pass through two convolutional (Conv) layers with GeLU activation. Subsequently, a skip connection is implemented to facilitate the information flow throughout the entire network. The global context former encoder consists of a $3\times3$ DWConv and a $1\times1$ Conv to learn spatial information and encode local representations. The resulting feature map is fed into an efficient additive attention block to learn contextual information at each scale of the input size. Finally, the output global features $\bm{V}_{G}$ is processed via a linear block, comprising two $1\times1$ Conv layers, a normalization layer, and GeLU activation. The Ho-Pair Encoder is defined as: 
\begin{align}
    \bm{{V_F^{'}}} = Conv_1(Conv_1,DWConv_3(BN(\bm{V_F}))+\bm{V_F},
\end{align}
\begin{align}
    \bm{{V_F^{'}}} = Conv_1(Conv_1,DWConv_3(BN(\bm{V_F^{'}})),
\end{align}
\begin{align}
    \bm{{V_G}} =Linear(QK({\bm{V_F^{'}}})+{\bm{V_F^{'}}}).
\end{align}

 \begin{figure}[t]
    \centering
    \includegraphics[width=1\linewidth]{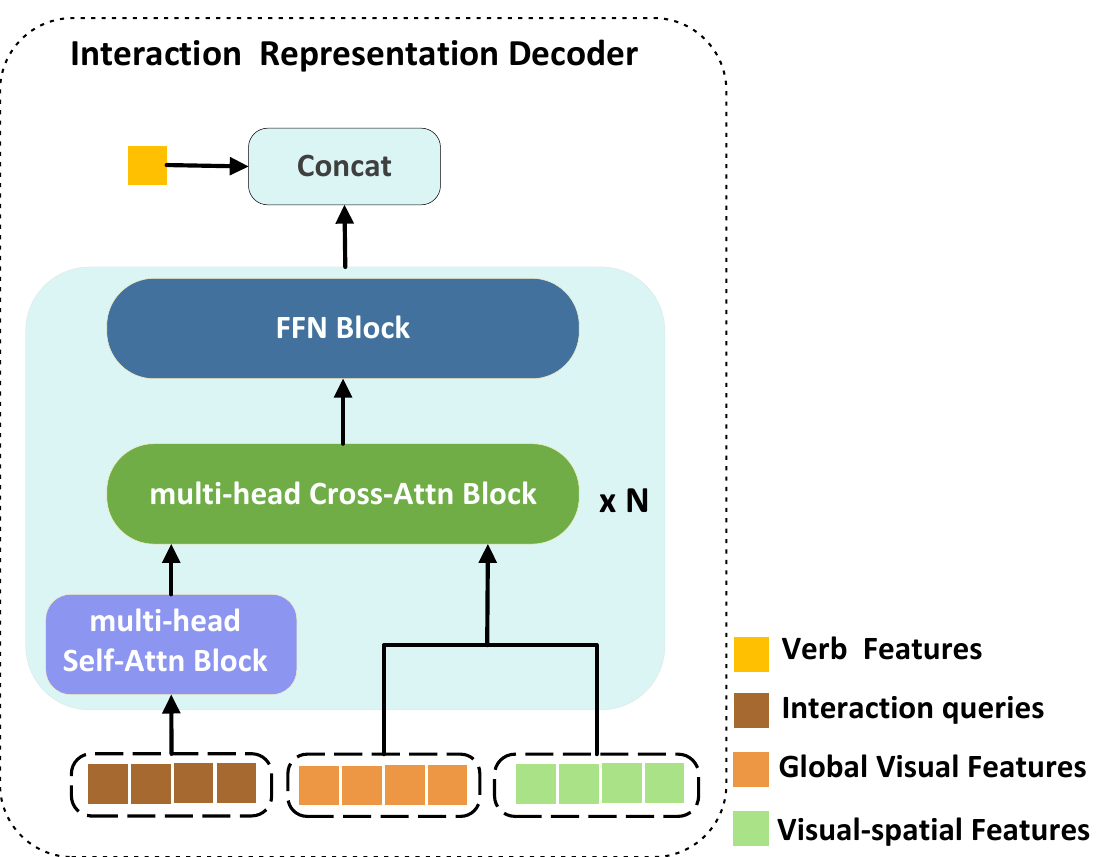}
    \caption{Structure of interaction representation decoder. Interaction queries $\bm{Q}_{inter}$, visual-spatial features $\bm{V}_{sp}$, and global visual features$ \bm{V}_{G}$ are passed through the multi-head cross-attention block before being fed into the multi-head self-attention block. Then, the outputs are concatenated with the verb features $\bm{V}_{verb}$ before being fed into the feed-forward network.}
    \label{fig4}
\end{figure} 

\subsection {Verb Feature Learning}\label{sec-3-3}
Inspired by the Query2Label~\cite{liu2021query2label}, we establish a novel approach termed learnable verb-specific query, where $\bm{{Q}_v}{\in}\mathbb{R}^{N\times{A}}$, A is the number of interaction categories. $\bm{{Q}_v}$ differs from traditional transformer queries by establishing a one-to-one correspondence between each query and a particular instance throughout the training and inference phases.   Such queries interact with global visual feature $\bm{{V}_g}$ through a verb extraction decoder and become verb-specific queries. We design a self-attention and multi-head attention combination module and a feed-forward network (FFN) layer as verb extraction decoder, consisting of two layers. A set of verb queries $\bm{{Q}_v}$ aggregates the global feature information $\bm{{V}_G}$ through verb extraction decoder and is updated to ${\bm{V}_{verb}}$. In this way, the queries are learned to capture the verb \textit{priors} and become good feature representations for these interactions. namely:
\begin{align}
    \hat{\bm{V}_G}=\text{selfAtten}(\hat{\bm{V}_G}),
\end{align}
\begin{align}
    \bm{V}_{verb} =\text{MultiheadAttn}(\hat{\bm{V}_G},\bm{Q}_V),
\end{align}
\begin{align}
    \bm{V}_{verb} = \text{FFN}(\bm{V}_{verb}).
\end{align}

\subsection{Interaction Semantic Representation (ISR)}\label{sec-3-4}
Conventional HOI detection uses an Interaction Classification Head to predict the confidence of each action for each pair of interactions and to judge the interaction. If the training sample for one of the actions is assumed to be too few or non-existent, the prediction for that action will be very inaccurate. This situation is more obvious in the zero-shot learning tasks, which is one of the reasons why traditional methods fail in the field of zero-shot learning. HOI detector maps visual features and tag-generated text features into the same space through CLIP, which performs quite well in the zero-shot learning tasks. We design the HOI labels and object labels are represented as ``A photo of a person [verb] a/an [object]" and ``A photo of a/an [object]" to assign different tokens to each HOI instance. To fully explore the CLIP knowledge, we propose to retrieve the text embeddings from the CLIP to better align them with the prior knowledge in the classifier weights.

\textbf{Interaction Representation Decoder. }We introduce an interaction semantic representation (ISR) module to extract the interactive representations of human-object interaction pairs. 
First, 
We add a learnable position guided embedding $\bm{P}{\in}\mathbb{R}^{N \times C}$ for human queries and object queries, \textit{viz.}, $\bm{Q}_h{\in}\mathbb{R}^{N \times C}$ and $\bm{Q}_o{\in}\mathbb{R}^{N \times C}$ at the same position as  interaction pairs. We compute the interaction queries $\bm{Q}_{inter}{\in}\mathbb{R}^{N \times C}$ by taking the average of the concatenation of $\bm{Q}_h$ and $\bm{Q}_o$. That is:
%The concatenate between these sets is subsequently employed as interaction queries $Q_{inter}{\in}R^{N \times C}$. 
\begin{align}
    \bm{Q}_{inter} = \text{Cat}(\bm{Q}_h+\bm{P},\bm{Q}_o+\bm{P})/2.
\end{align}

To guide interaction queries $\bm{Q}_{inter}$ to explore informative regions in both $\bm{V}_{G}$ and $\bm{V}_{sp}$, we design an interaction representation decoder with multiple cross-attention, as shown in Figure~\ref{fig4}. Each decoder consists of an attention block, a self-attention block, and a forward feedback network. $\bm{Q}_{inter}$ is first input to the self-attention block and the corresponding output is fed into the cross-attention mechanism with $\bm{V}_{G}$ and $\bm{V}_{sp}$. Subsequently, the final output is:
\begin{align}
    \bm{Q}_{inter} = \text{MHSA}(\bm{Q}_{inter}),
\end{align}
\begin{align}
    \bm{Q}_{inter}^{'} = \text{MHCA}(\bm{Q}_{inter},\bm{V}_{sp}),
\end{align}
\begin{align}
    \bm{Q}_{inter^{''}} = \text{MHCA}(\bm{Q}_{inter},\bm{V}_{G}),
\end{align}
\begin{align}
     \bm{Q}_{inter}=\text{FFN}(\bm{Q}_{inter}^{'}+\bm{Q}_{inter}^{''}),
\end{align}
where MHSA denotes a multi-head self-attention operation and MHCA denotes a multi-head cross-attention operation. Since, the obtained $\bm{Q}_{inter}{\in}\mathbb{R}^{N_q \times C}$ integrates the knowledge of CLIP and visual features, enabling the detection of fine-grained HOI.

By leveraging object and human information from the instance Interactor, we can concatenate interaction representations from the spatial feature map of CLIP and visual features from the detector to efficiently retrieve corresponding interaction representations and achieve strong generalization capabilities.

\textbf{Verb Predictor via Knowledge Retrieval. }To align verb features \bm{$V_{verb}$} with $\bm{Q}_{inter}$, we employ projection operation to map both $\bm{Q}_{inter}$ and $\bm{V}_{verb}$ into the CLIP feature space. After projection, multi-layer perceptron (MLP) is applied to extract $\bm{D}_{verb}$ from $\bm{Q}_{inter}$ and $\bm{C}_{verb}$ from $\bm{V}_{verb}$ from $\bm{Q}_{inter}$ and $\bm{V}_{verb}$. Finally, the verb score is obtained as:

\begin{align}
     \bm{Q}_{inter}=\text{Proj}(\bm{Q}_{inter}), \bm{V}_{verb}=\text{Proj}(\bm{V}_{verb}),
\end{align}
\begin{align}
     \bm{C}_{verb}=\text{MLP}(\bm{V}_{verb}) , \bm{D}_{verb}=\text{MLP}(\bm{Q}_{inter}),
\end{align}
\begin{align}
     \bm{C}_{verb}=\text{Cat}(\bm{C}_{verb},\bm{D}_{verb}),
\end{align}
\begin{align}
     \bm{S}_{verb}=\bm{C}_{verb}\bm{W^T}_v,
\end{align}
where the verb score is computed by the cosine similarity between the verb features and the text weight of the verb representations. As well, a reconstruction loss function that quantifies the dissimilarity between features. That is:
\begin{align}
     \bm{L}_{re}=L_1(\bm{Q}_{inter},\bm{V}_{sp}),
\end{align}
where $L1$ loss is used to minimize the distance between features and visual embeddings.
\begin{table}[h]
    \centering
     \caption{Performance comparison for zero-shot HOI detection on HICO-DET. RF-UC indicates rare first setting, and NF-UC represents non-rare first unseen combination setting. UC, UO, and UV denote unseen composition, unseen object, and unseen verb settings, respectively.}
    \resizebox{\linewidth}{!}{ 
    \begin{tabular}{lllll}
        \toprule  
        Method  & Type & Full & Seen &Unseen \\
        \midrule
        VCL~\cite{hou2020visual} &RF-UC &21.43 &24.28 &10.06 \\
        ATL~\cite{hou2021affordance} &RF-UC &21.57 &24.67 &9.18 \\
        FCL~\cite{hou2021detecting}  &RF-UC &22.01 &24.23 &13.16 \\
        SCL~\cite{hou2022discovering} &RF-UC &28.08 &30.39 &19.07 \\
        RLIP~\cite{yuan2022rlip} &RF-UC &30.52 &33.35 &19.19 \\
        EoID~\cite{wu2023end} &RF-UC &29.52 &31.39 &22.04 \\
        GEN-VLKT~\cite{9878997}  &RF-UC &30.56 &32.91 &21.36\\
        HOICLIP~\cite{10204103} &RF-UC &32.99 &34.85 &25.53\\
        \rowcolor[gray]{0.9} \textbf{KI2HOI} & RF-UC &\textbf{34.10 }&\textbf{35.79} &\textbf{26.33} \\
        \midrule
        VCL~\cite{hou2020visual} &NF-UC &16.22 &18.52 &18.06 \\
        ATL~\cite{hou2021affordance} &NF-UC &18.25 &18.78 &18.67\\
        FCL~\cite{hou2021detecting}  &NF-UC &18.66 &19.55 &19.37\\
        SCL~\cite{hou2022discovering} &NF-UC &24.34 &25.00 &21.73 \\
        RLIP~\cite{yuan2022rlip} &NF-UC &26.19 &27.67 &20.27 \\
        GEN-VLKT~\cite{9878997}  &NF-UC &23.71 &23.38 &25.05\\
        EoID~\cite{wu2023end} &NF-UC &26.69 &26.66 &26.77\\
        HOICLIP~\cite{10204103} &NF-UC &27.75 &28.10 &26.39\\
        \rowcolor[gray]{0.9} \textbf{KI2HOI} &NF-UC &\textbf{27.77} &\textbf{28.31} &\textbf{28.89}\\
        \bottomrule
        FG~\cite{bansal2020detecting} & UC &12.26 &12.60 &10.93\\
        ConsNet~\cite{liu2020consnet} & UC & 19.81 &20.51 &16.99\\
        EoID~\cite{wu2023end}   & UC  & 28.91 &30.39 &23.01 \\ 
        HOICLIP~\cite{10204103} & UC &32.99 &34.85 &25.53 \\
        \rowcolor[gray]{0.9} \textbf{KI2HOI} & UC &\textbf{34.56} &\textbf{35.76} &\textbf{27.43} \\
       \bottomrule
        FCL~\cite{hou2021detecting}  &UO &19.87 &20.74 &15.54\\
        ATL~\cite{hou2021affordance} &UO &20.47 &21.54 &15.11\\
        GEN-VLKT~\cite{9878997}  &UO &25.63 &28.92 &10.51\\
        HOICLIP~\cite{10204103} &UO &28.53 &30.99 &16.20\\
        \rowcolor[gray]{0.9} \textbf{KI2HOI} &UO &\textbf{28.84} &\textbf{31.70} &\textbf{16.50} \\
      \bottomrule
        ConsNet~\cite{liu2020consnet} &UV &19.04 &20.02 &14.12\\
        GEN-VLKT~\cite{9878997}  &UV &28.74 &30.23 &20.96 \\
        EoID~\cite{wu2023end}  &UV &29.61 &30.73 &22.71 \\
        HOICLIP~\cite{10204103} &UV &31.09 &32.19 &24.30\\
         \rowcolor[gray]{0.9}  \textbf{KI2HOI} &UV &\textbf{31.85} &\textbf{32.95} &\textbf{25.20} \\
    \bottomrule
    \end{tabular}}
   
    \label{tab:Zero-Shot}
\end{table}

\begin{table*}[h]
 \centering
  \caption{Comparison with state-of-the-art methods on HICODET and V-COCO. All methods employ the ResNet-50 backbone network.}
  \begin{tabular*}{0.95\textwidth}{ccccccccccc}
    \toprule
    \multirow{2}{*}{Method} & \multirow{2}{*}{Detector} & \multirow{2}{*}{Backbone} & \multicolumn{3}{c}{mAP Default} & \multicolumn{3}{c}{mAP Know Object} & \multicolumn{2}{c}{V-COCO} \\
    \cmidrule{4-6} \cmidrule{7-9} \cmidrule{10-11}
    & & & Non-Rare & Full & Rare & Non-Rare & Full & Rare & $AP_{role}^{S1}$ & $AP_{role}^{S2}$ \\
    \midrule
         IDN~\cite{li2020hoi} &COCO &ResNet-50 & 23.36 & 22.47 & 23.63 & 26.43 & 25.01 & 26.85 &53.3 & 60.3 \\
         HOTR~\cite{kim2021hotr} &HICO-Det &ResNet-50 & 25.10 & 17.34 & 27.42 & -     & -     & -    &55.2 &64.4 \\
         ATL~\cite{hou2021affordance}  &COCO &ResNet-50 & 28.53 & 21.64 & 30.59 & 31.18 & 24.15 & 33.29 &- &-\\
     
         QPIC~\cite{tamura2021qpic} &HICO-Det   &ResNet-50 & 29.07 & 21.85 & 31.23 & 31.68 & 24.14 & 33.93 &58.8 &61.0\\
         FCL~\cite{hou2021detecting} & COCO &ResNet-50  & 29.12 & 23.67 & 30.75 & 31.31 & 25.62 & 33.02 &52.4 &-\\
         SCG~\cite{zhang2021spatially}  &COCO & ResNet-50-FPN  & 31.33 & 24.72 & 33.31 & 34.37 & 27.18 & 36.52 &54.2 &60.9\\
         UPT~\cite{zhang2022efficient} &COCO &ResNet-50  & 31.66 & 25.94 & 33.36 & 35.05 & 29.27 & 36.77 &59.0 &64.5\\
         CDN~\cite{zhang2021mining}  &HICO-DET  &ResNet-50  & 31.78 & 27.55 & 33.05 & 34.53 & 29.73 &35.96 &62.3 &64.4\\
         Iwin~\cite{tu2022iwin} &HICO-Det &ResNet-50-FPN  & 32.03 & 27.62 & 34.14 & 35.17 & 28.79 & 35.91 & 60.5 &-\\
         ~\cite{liu2022interactiveness} &COCO &ResNet-50 & 33.51 & 30.30 & 34.46 & 36.28 & 33.16 & 37.21 &63.0 &65.2\\
         GEN-VLKT~\cite{9878997} &HICO-Det &ResNet-50+ViT-B & 33.75 & 29.25 & 35.10 & 36.78 & 32.75 & 37.99 &62.4 &64.4\\
         ADA-CM~\cite{lei2023efficient} &COCO &ResNet-50+ViT-B  &33.80 &31.72 &34.42&-&-&- &56.1 &61.5 \\
         OpenCat~\cite{zheng2023open} &-  &ResNet-50 &32.68 &28.42 &33.75 &-& -& - &61.9 &63.2\\
         HOICLIP~\cite{10204103} &HICO-Det &ResNet-50+ViT-B &\textbf{34.69} & 31.12 & 35.74 & 37.61 & 34.47 & 38.54 &63.5 &64.8\\
        \rowcolor[gray]{0.9} \textbf{KI2HOI} &HICO-Det &ResNet-50+ViT-B  &34.20 &\textbf{32.26}  & \textbf{36.10} &\textbf{37.85}  &\textbf{35.89}  &\textbf{38.78} &\textbf{63.9}&\textbf{65.0} \\ 
        \toprule 
    \end{tabular*}
    \label{tab:hico}
\end{table*}
\begin{table}[t]
    \centering
    \caption{Robustness to different distribution data.}
    \begin{tabular}{l|cccc}
         \hline  
          &\multicolumn{4}{c}{\textbf{Non-Rare HICO-DET}} \\
        \hline  
         Percentage &100$\%$ &50$\%$ &25$\%$ &15$\%$  \\
         \hline
         GEN-VLKT   &33.75  &26.55  &22.14  &20.40   \\
         \rowcolor[gray]{0.9} \textbf{KI2HOI}    &34.20 &31.25 &30.06 &27.20   \\
         mAP gain($\%$) &+1.3 &+19.2 &+35.77 &+33.3  \\
      \hline  
          &\multicolumn{4}{c}{\textbf{Rare HICO-DET}} \\
      \hline  
         Percentage &100$\%$ &50$\%$ &25$\%$ &15$\%$ \\
      \hline  
         GEN-VLKT  &29.25 &18.94 &14.04 &13.84 \\
         \rowcolor[gray]{0.9} \textbf{KI2HOI}   &36.10 &26.68 &25.05 &22.51  \\
         mAP gain($\%$)  &+23.41 &+40.86 &+78.41 &+38.51 \\
         \hline
    \end{tabular}
    \label{tab:data}
\end{table}
\begin{table}[h]
    \centering
     \caption{Network architecture analysis. Ablation studies are conducted on HICO-DET under the Unseen Verb (UV) setting.}
    \begin{tabular}{lccc}
         \toprule  
            Method  & Full & Seen &Unseen \\
        \midrule
        Baseline &28.20 &30.49 &9.57 \\
        +CLIP  &30.45 &31.65 &23.37 \\
        +Ho-Pair Encoder  &30.99 & 32.02 &23.96 \\
        \rowcolor[gray]{0.9}+Verb Feature Learning   &\textbf{31.85} &\textbf{32.95} &\textbf{25.20} \\
         \toprule
    \end{tabular}
    \label{tab:Architecture}
\end{table}
\begin{table}[t]
    \centering
    \caption{Reconstruction loss setting.}
    \begin{tabular}{cc|ccc}
         \toprule  
            $L_1$  &$L_2$ & Full & Rare &Non-Rare \\
        \midrule
         -  &-    &33.31 &31.83 &35.05\\
          \rowcolor[gray]{0.9} \checkmark &-  &34.20 &32.26 &36.10\\
        - &\checkmark &34.06 &30.65 &35.08\\
        \checkmark &\checkmark &34.08 &30.18 &35.24\\
         \toprule
    \end{tabular}
    \label{tab:LOSS}
\end{table}
\begin{table}[t]
    \centering
    \caption{The impact of different verb extraction decoder layer numbers.}
    \begin{tabular}{c|ccc}
         \toprule  
            Layers  & Full & Seen &Unseen \\
        \midrule
        \rowcolor[gray]{0.9} 1 &27.77 &28.31 &28.89\\
            2   &26.30 &26.70 &24.71\\
            3   &26.53 &26.70 &25.07\\
         \toprule
    \end{tabular}
    \label{tab:layers}
\end{table}

\subsection{Training and Inference}\label{sec-3-5}
\textbf{Training.} We employ the Hungarian algorithm
~\cite{wu2023end,9878997,10204103} for bipartite matching between predictions and ground truths. The matching cost consists of human bounding box regression loss $\bm{L}_{bh}$, object bounding box regression loss $\bm{L}_{bo}$, object classification loss $\bm{L}_o$, interaction-over-union loss $\bm{L}_u$, and interaction classification loss \bm{$L_i$}. Combined with the reconstruction loss function, the final loss function is as follows:
\begin{align}
\bm{L}=\lambda_{\alpha}\bm{L}_b+\lambda_{\beta}\bm{L}_u+\lambda_{\delta}\bm{L}_i+\lambda_{re}\bm{L}_{re},
\end{align}
where $\lambda_{re}$ is weighting parameter, $\lambda_{\alpha}$, $\lambda_{\beta}$, $\lambda_{\delta}$ are hyper-parameters for adjusting the weights of all losses, $\bm{L}_b = \bm{L}_{bh}+\bm{L}_{bo}$.

\textbf{Inference.} Reconstruction loss is only used for training, $\bm{S}_h\in[0,1]^{N}$, $\bm{S}_o\in[0,1]^{N}$. In inference, the final score $\bm{S}_{final}\in[0,1]^{N}$ is summed by $\bm{S}_h$, $\bm{S}_O$ and $\bm{S}_{verb}$.
\begin{align}
    \bm{S}_{final}^{i}=\bm{S}_h+\bm{S}_o+\bm{S}_{verb} , i\in[1,c].
\end{align}

\begin{figure*}[h]
    \centering   
    \subfigure{   
        \includegraphics[width=0.14\textwidth]{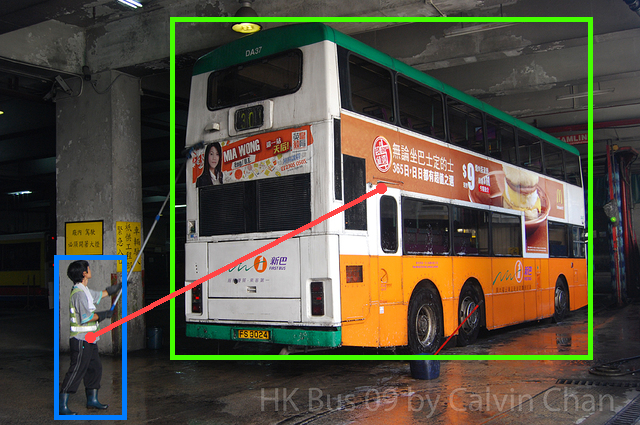}  
    }   
    \subfigure{ 
        \includegraphics[width=0.14\textwidth]{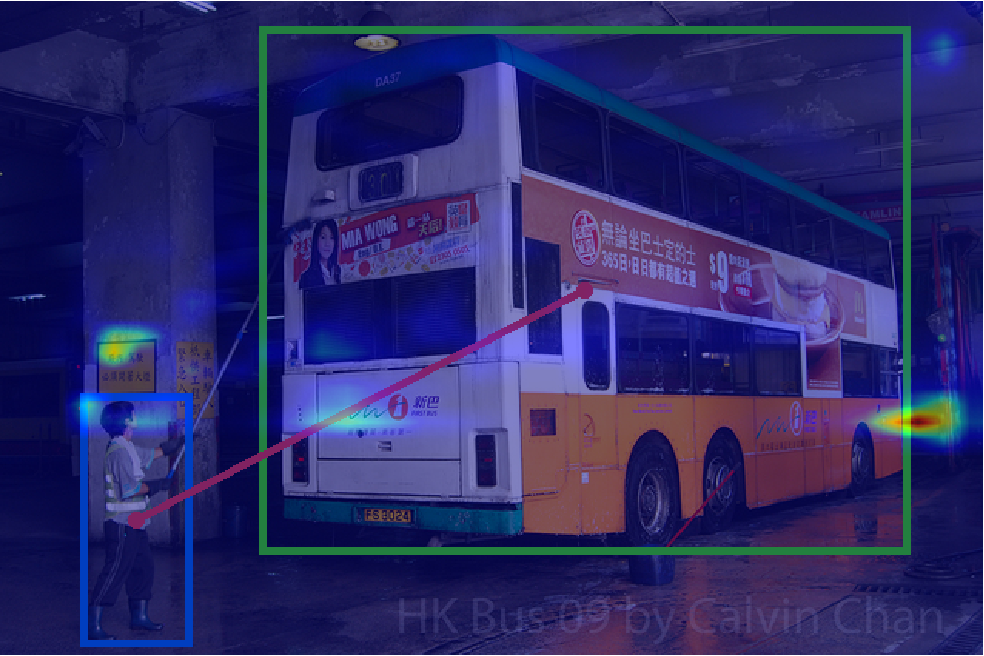}  
    } 
    \subfigure{   
        \includegraphics[width=0.14\textwidth]{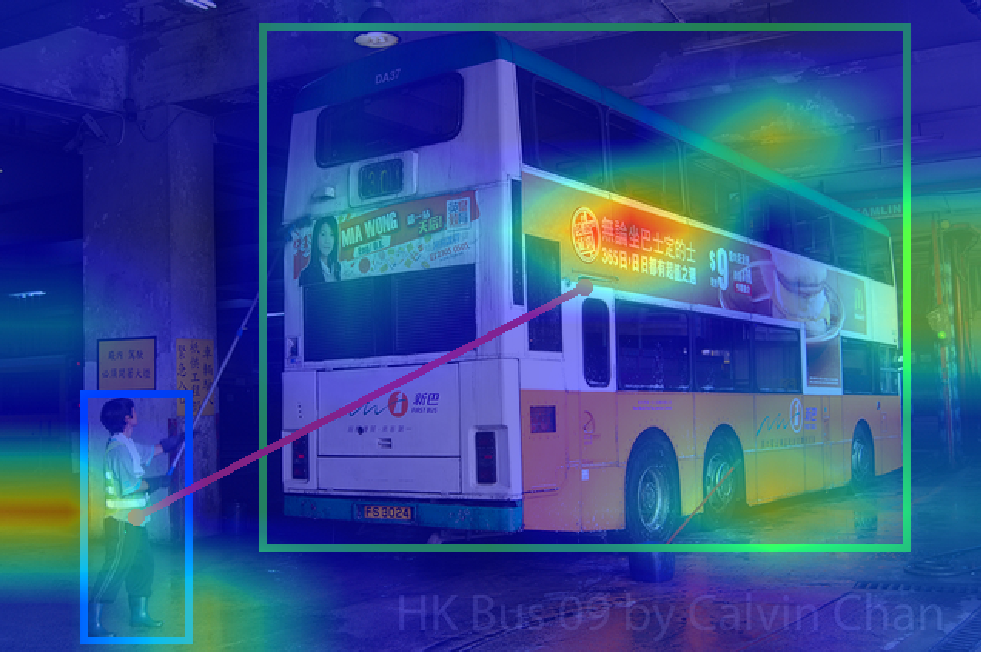}   
    }
    \subfigure{   
        \includegraphics[width=0.14\textwidth]{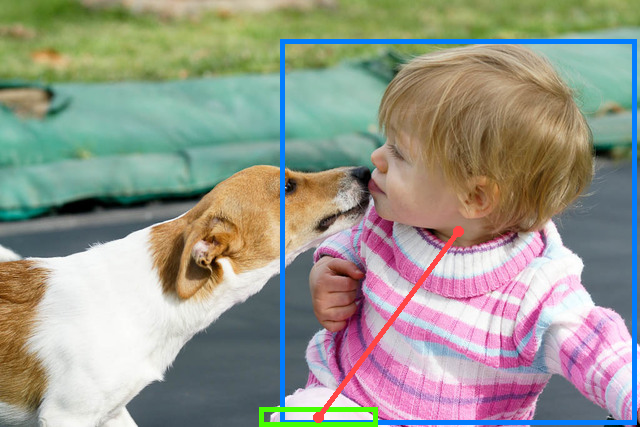}    
    }   
    \subfigure{   
        \includegraphics[width=0.14\textwidth]{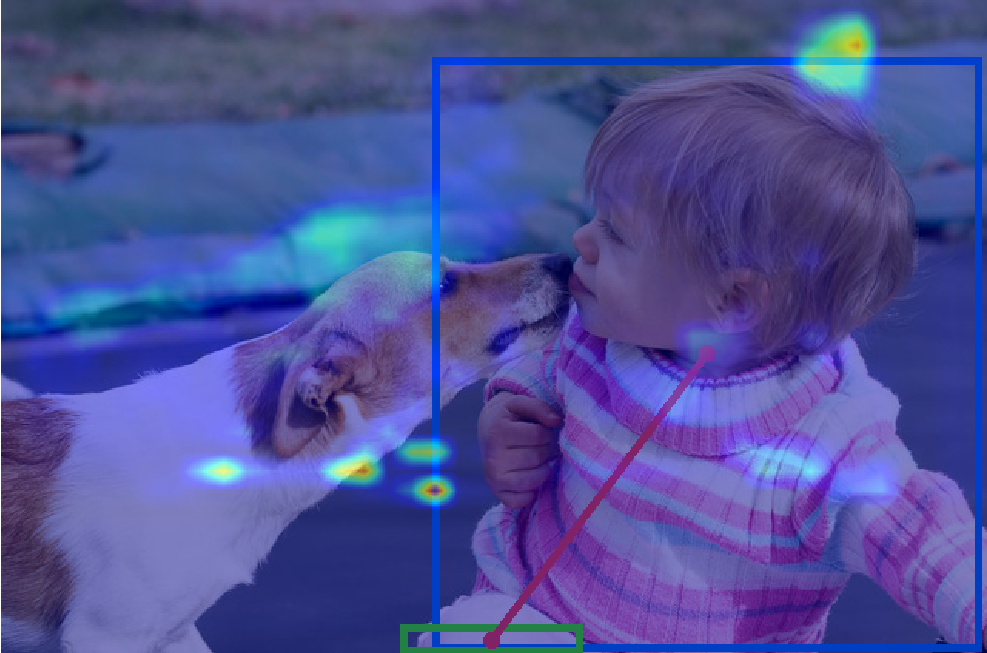}
    } 
    \subfigure{   
        \includegraphics[width=0.14\textwidth]{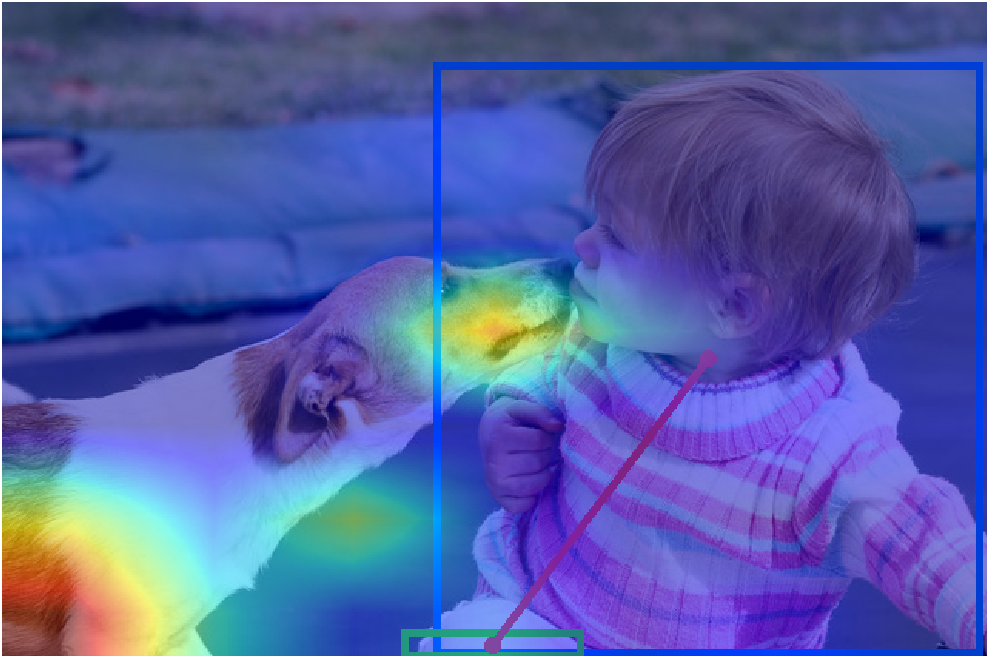}    
    }
      \subfigure{   
        \includegraphics[width=0.14\textwidth]{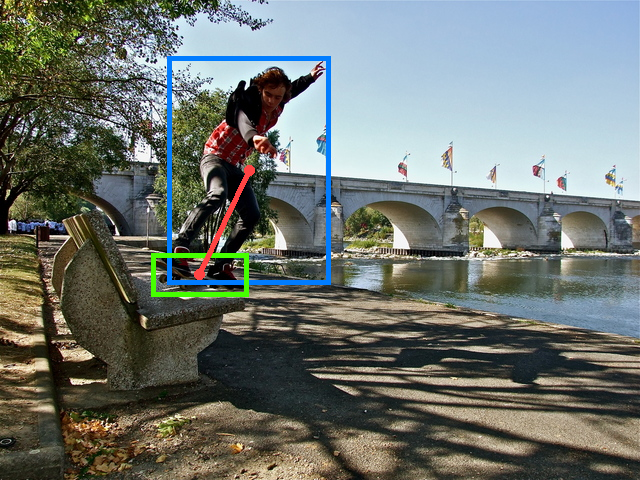}  
    }   
    \subfigure{ 
        \includegraphics[width=0.14\textwidth]{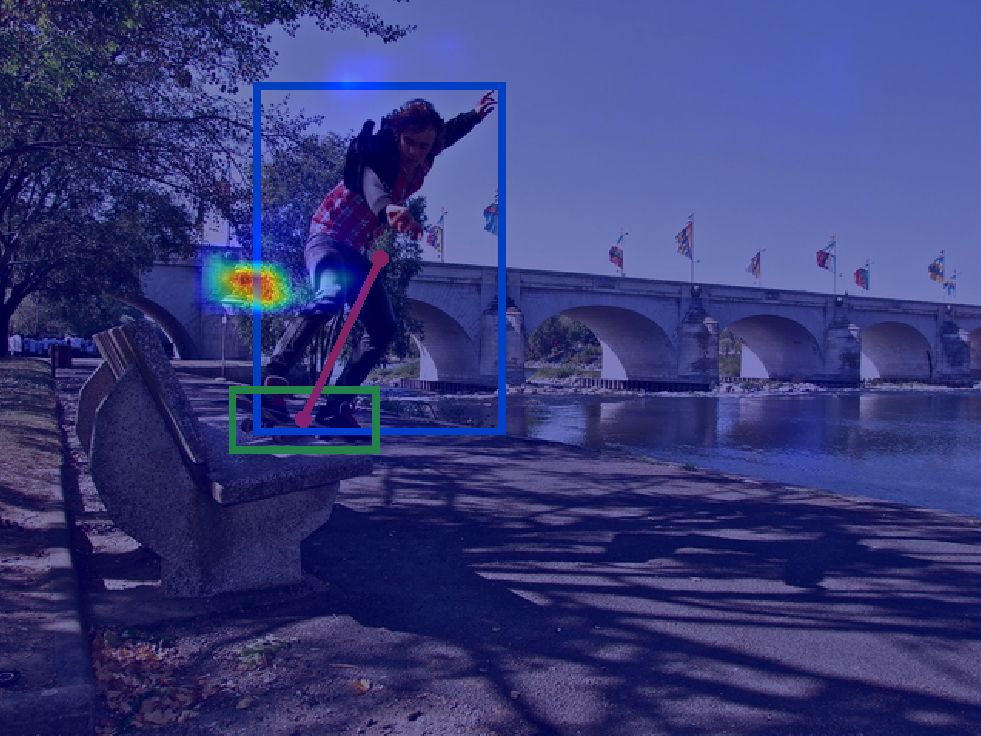}  
    } 
    \subfigure{   
        \includegraphics[width=0.14\textwidth]{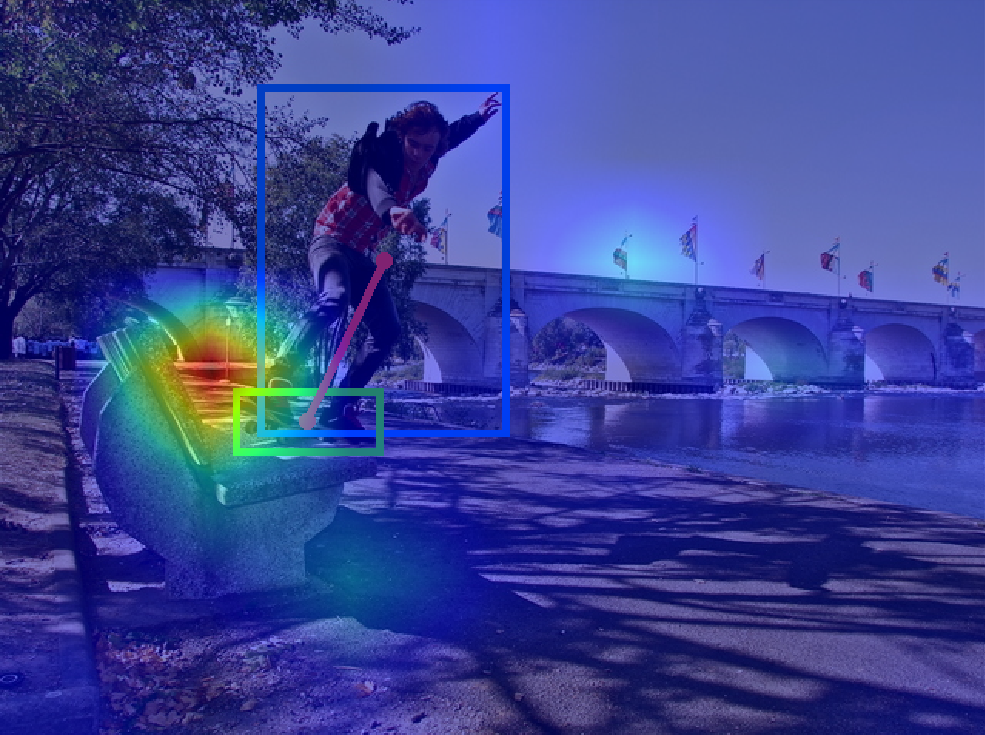}   
    }
    \subfigure{   
        \includegraphics[width=0.14\textwidth]{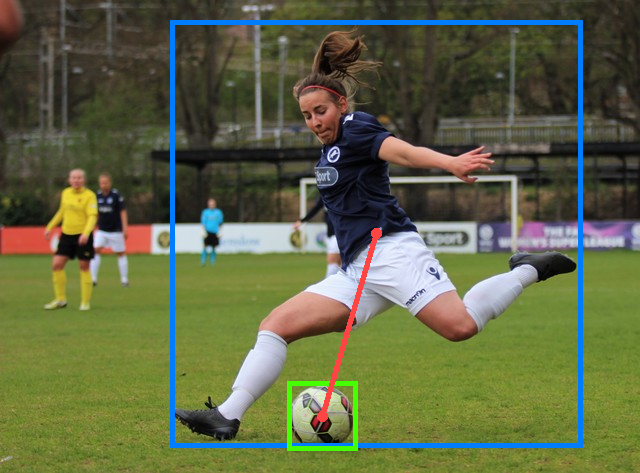}  
    }   
    \subfigure{ 
        \includegraphics[width=0.14\textwidth]{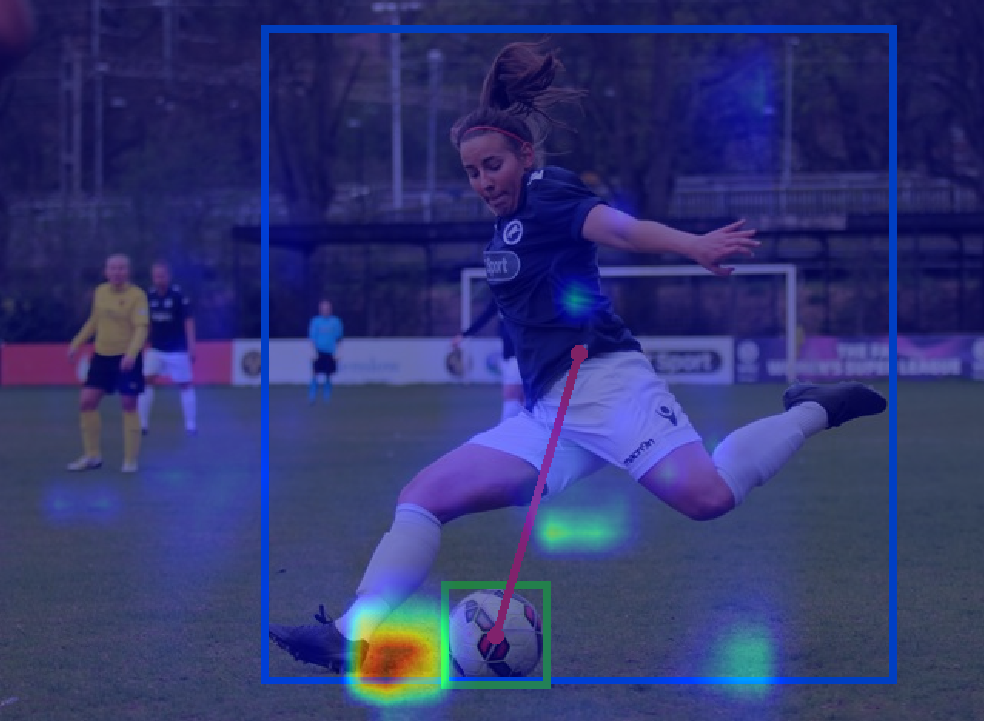}  
    } 
    \subfigure{   
        \includegraphics[width=0.14\textwidth]{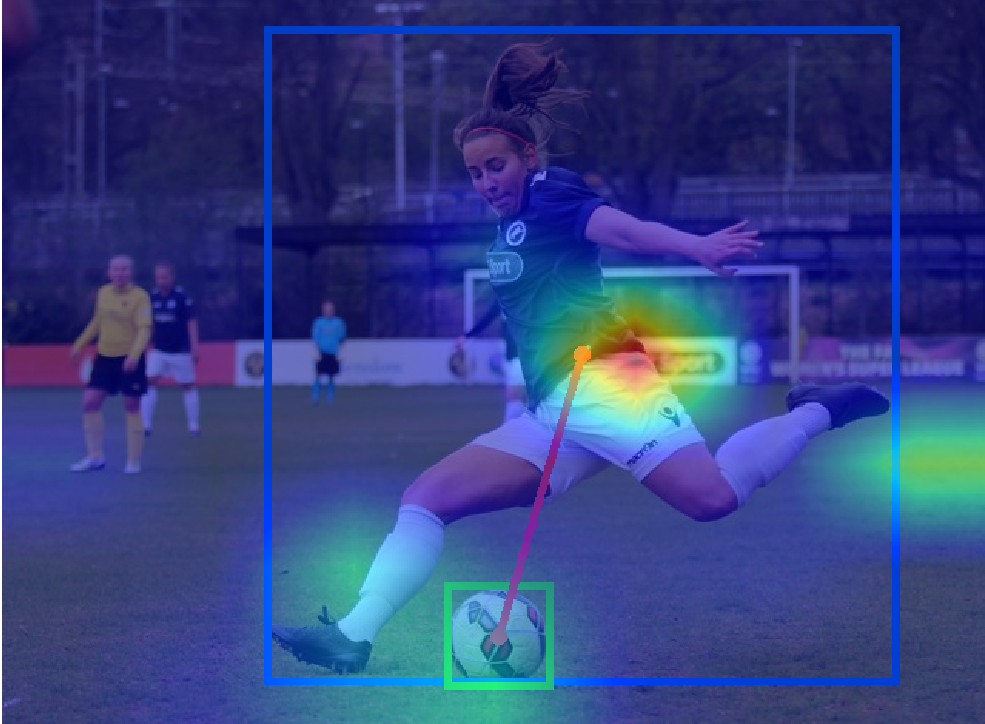}   
    }
     \subfigure{   
        \includegraphics[width=0.14\textwidth]{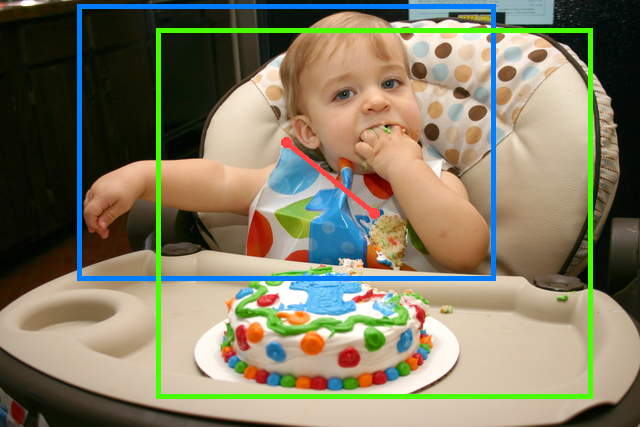}  
    }   
    \subfigure{ 
        \includegraphics[width=0.14\textwidth]{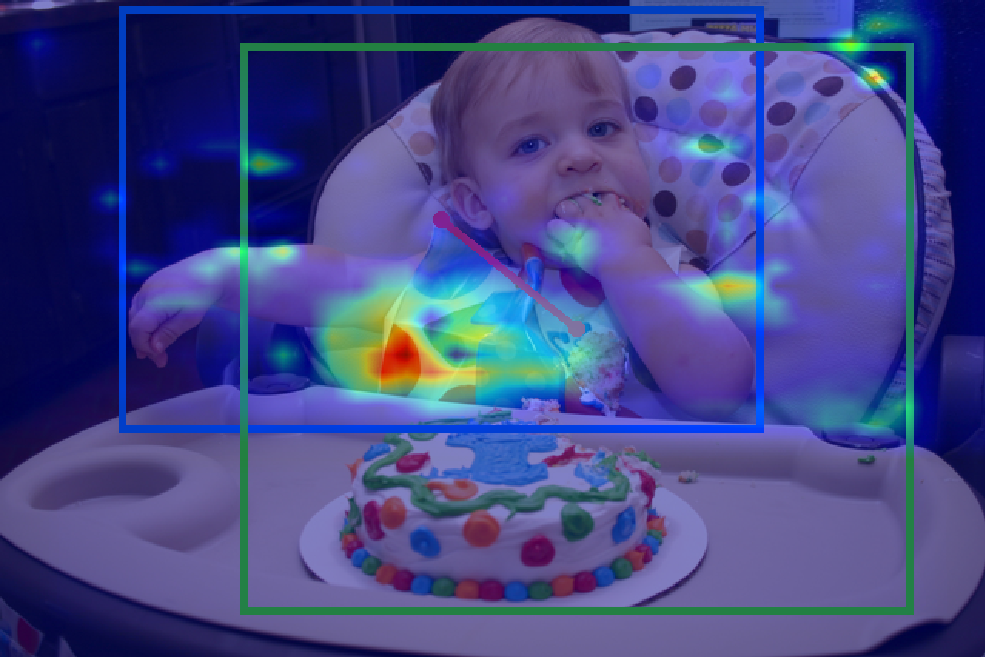}  
    } 
    \subfigure{   
        \includegraphics[width=0.14\textwidth]{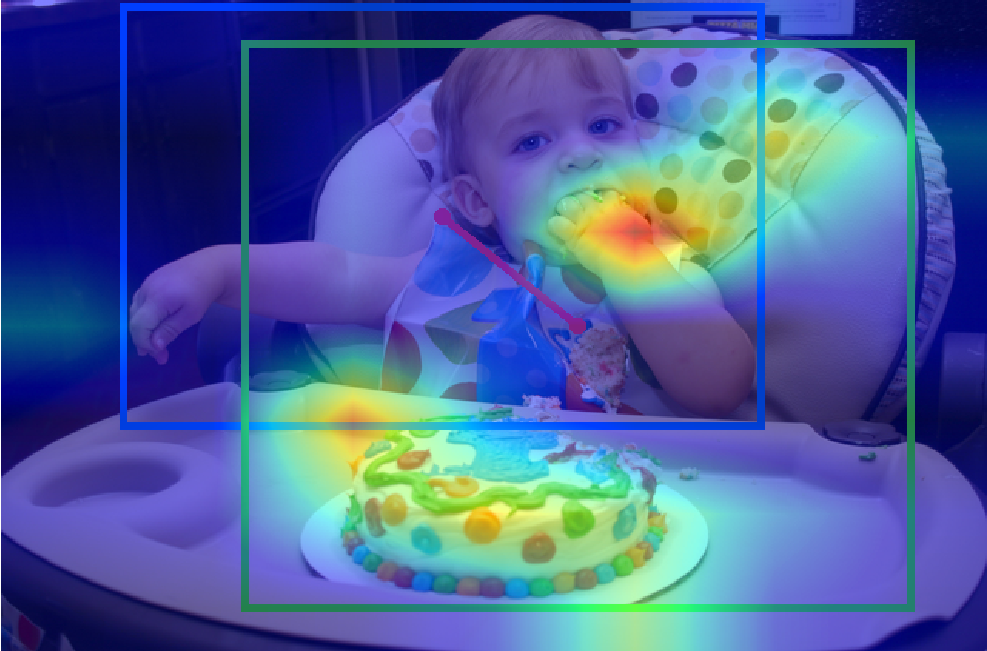}   
    }
    \subfigure{   
        \includegraphics[width=0.14\textwidth]{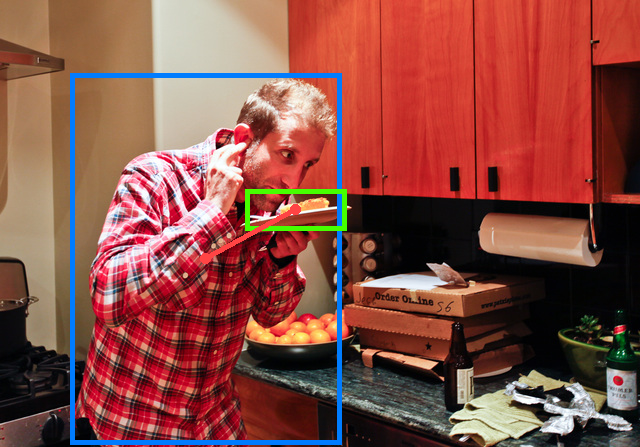}  
    }   
    \subfigure{ 
        \includegraphics[width=0.14\textwidth]{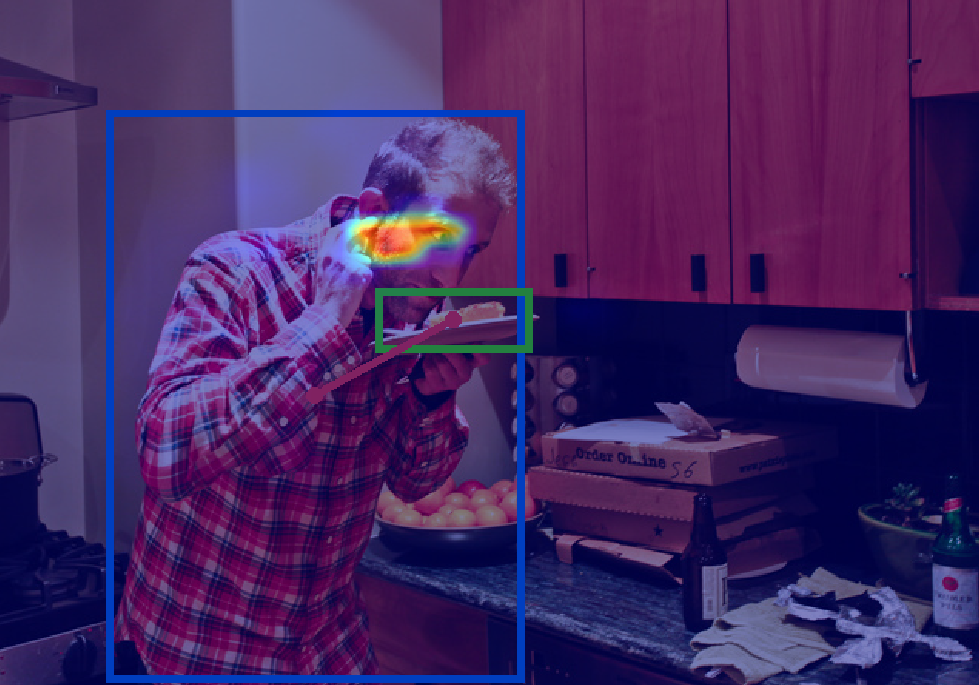}  
    } 
    \subfigure{   
        \includegraphics[width=0.14\textwidth]{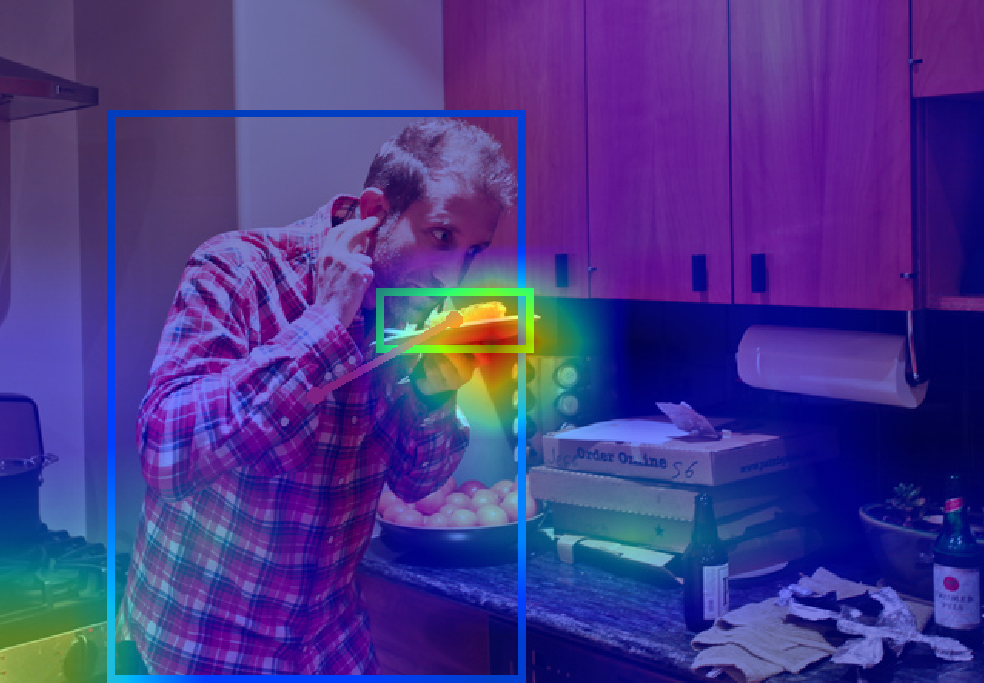}   
    }
  
  \subfigure{   
        \includegraphics[width=0.14\textwidth]{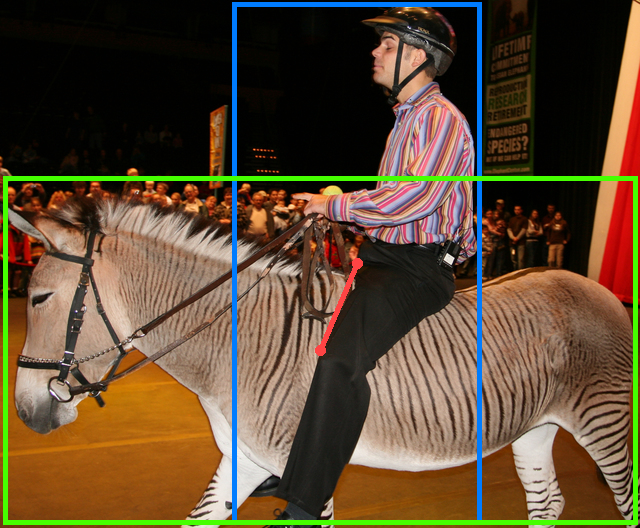}  
    }   
    \subfigure{ 
        \includegraphics[width=0.14\textwidth]{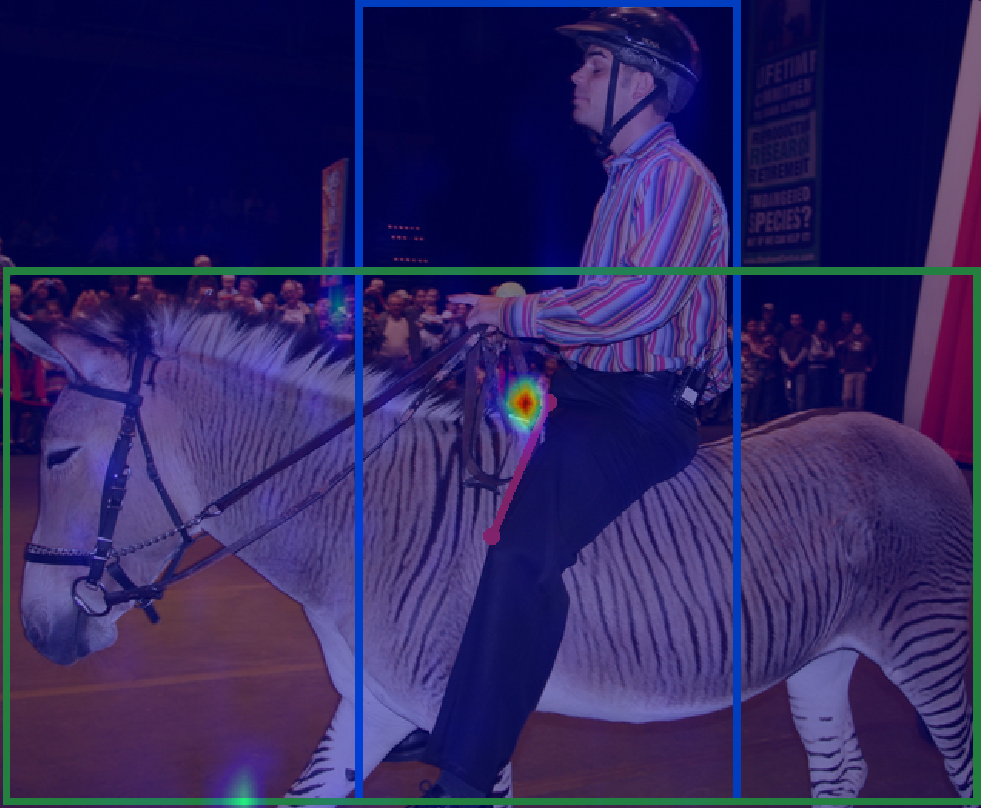}  
    } 
    \subfigure{   
        \includegraphics[width=0.14\textwidth]{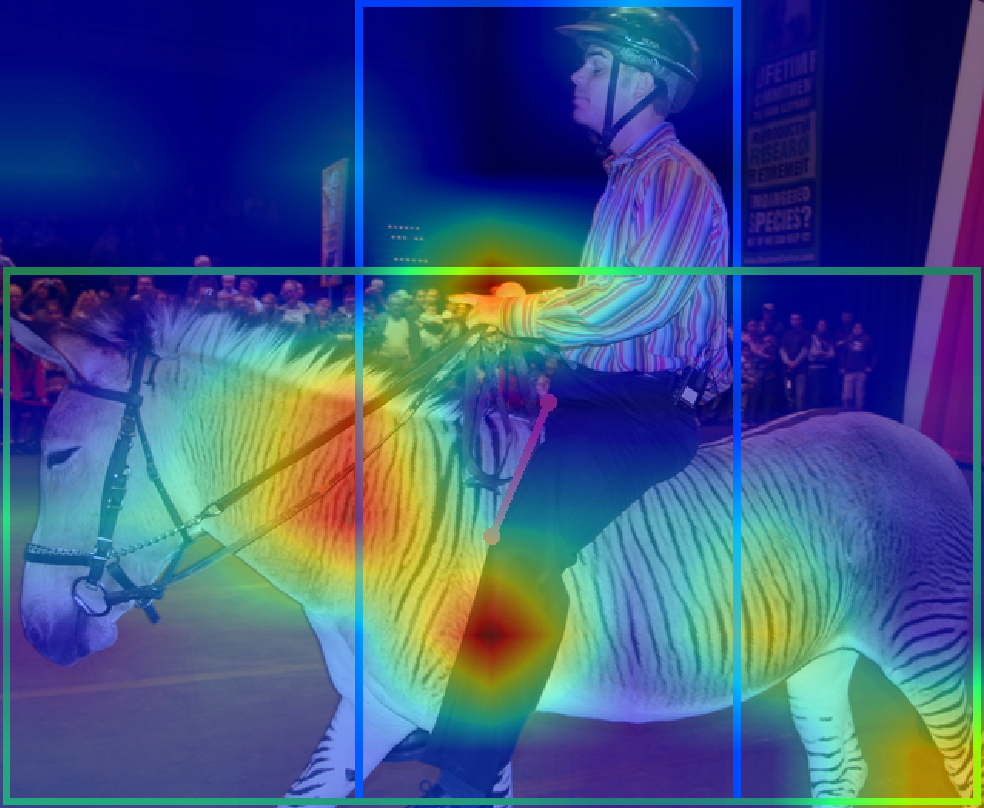}   
    }
      \subfigure{   
        \includegraphics[width=0.14\textwidth]{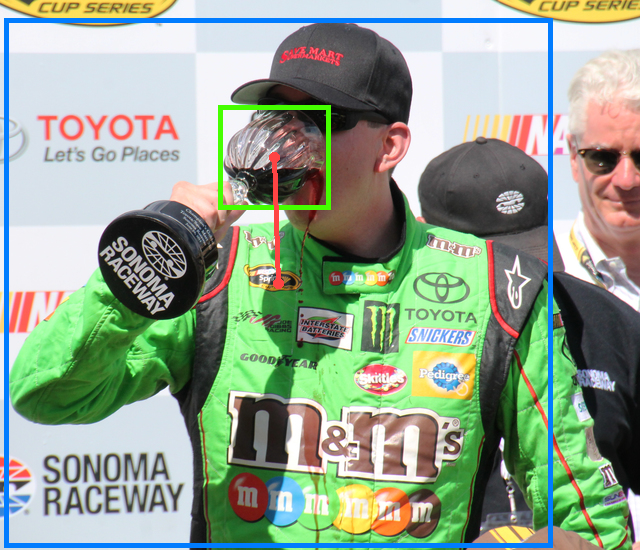}  
    }   
    \subfigure{ 
        \includegraphics[width=0.14\textwidth]{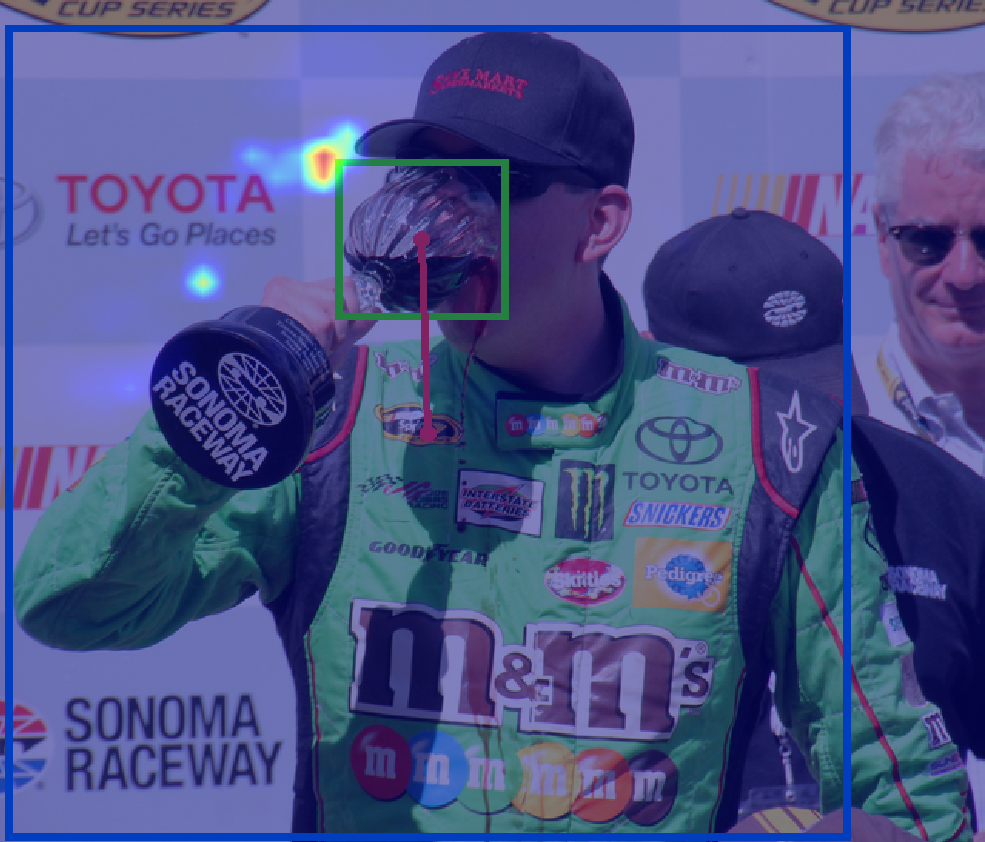}  
    } 
    \subfigure{   
        \includegraphics[width=0.14\textwidth]{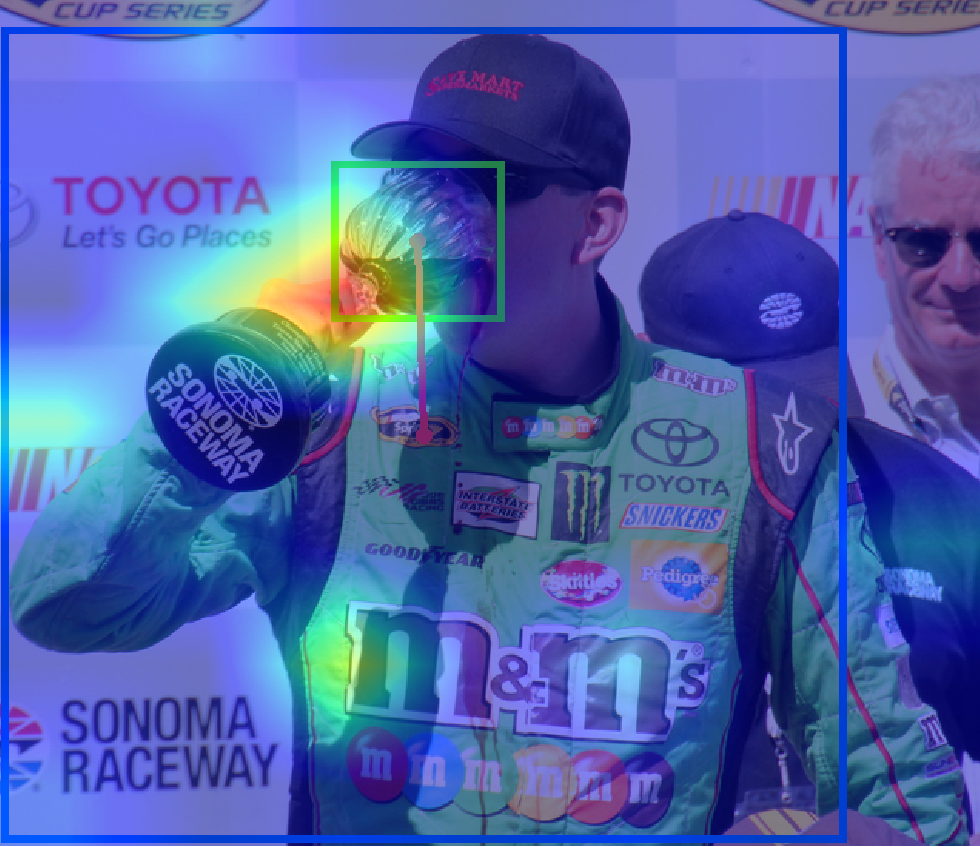}   
    }
    
    \caption{Visualization of the HOI detection results. From left to right, \textbf{column 1}: HOI prediction results; \textbf{column 2}: attention maps from Verb Extraction Decoder; \textbf{column 3}: attention maps from Interaction Representation Decoder. Images are sampled from the HICO-DET dataset in UV test set.}
    \label{Qualitative}
\end{figure*}

\section{EXPERIMENTS}
\subsection{Experimental Setup}
\begin{figure*}[h]
	\centering
    %第一行图片展示
     \subfigure{
        \rotatebox{90}{\scriptsize{One VS One}}
		\begin{minipage}[t]{0.2\textwidth}
		\centering
	   \includegraphics[width=0.9\linewidth]{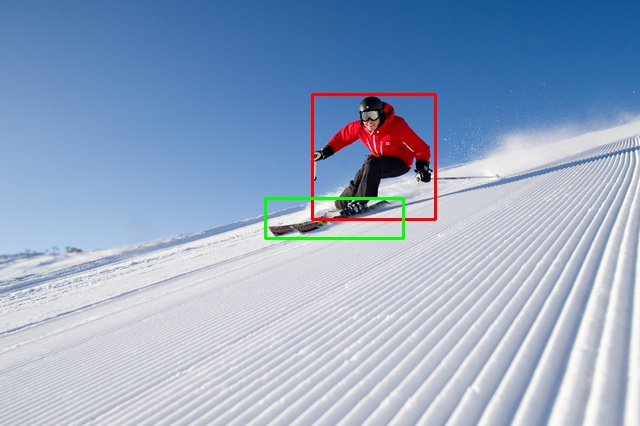}
             \centerline{\textcolor{red}{person} \textcolor{blue}{instr} \textcolor{green}{snowboard}}
   \end{minipage}
	}\vspace{-10pt}
	\subfigure{
		\begin{minipage}[t]{0.2\textwidth}
		\centering
		\includegraphics[width=0.9\linewidth]{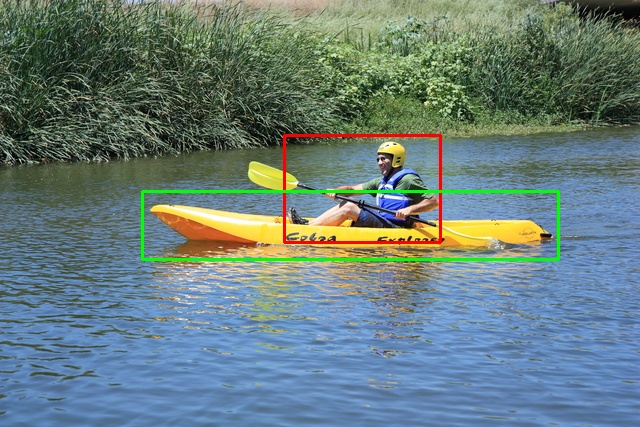}
        \centerline{\textcolor{red}{person} \textcolor{blue}{row} \textcolor{green}{boat} }
		\end{minipage}
	}
    \subfigure{
		\begin{minipage}[t]{0.2\textwidth}
		\centering
		\includegraphics[width=0.9\linewidth]{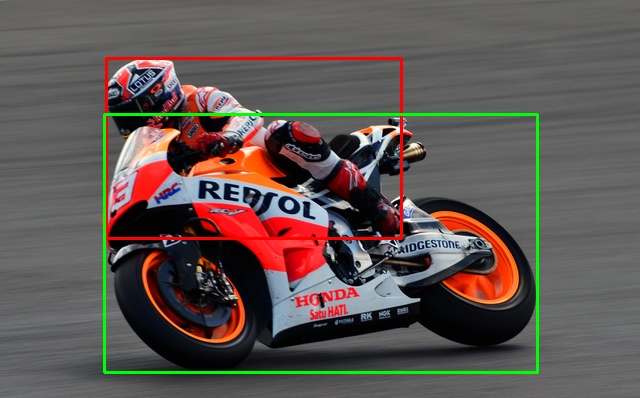}
        \centerline{\textcolor{red}{person} \textcolor{blue}{ride } \textcolor{green}{motorcycle} }
		\end{minipage}
	}
    \subfigure{
		\begin{minipage}[t]{0.2\textwidth}
		\centering
		\includegraphics[width=0.9\linewidth]{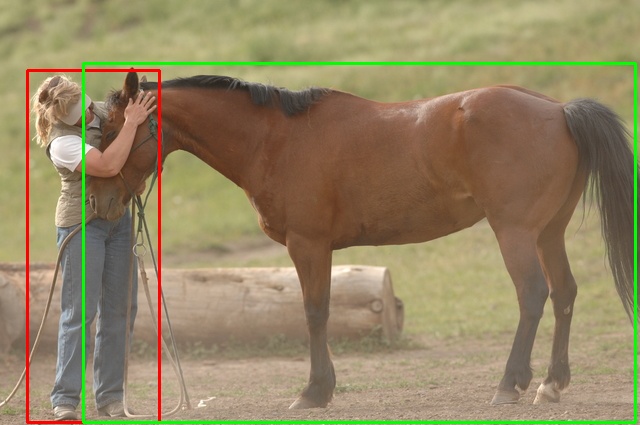}
        \centerline{\textcolor{red}{person} \textcolor{blue}{touch} \textcolor{green}{horse} }
		\end{minipage}
	}
 
	\subfigure{
		%左标题1
        \rotatebox{90}{\scriptsize{One VS Many}}
		\begin{minipage}[t]{0.2\textwidth}
			\centering
			\includegraphics[width=0.9\textwidth]{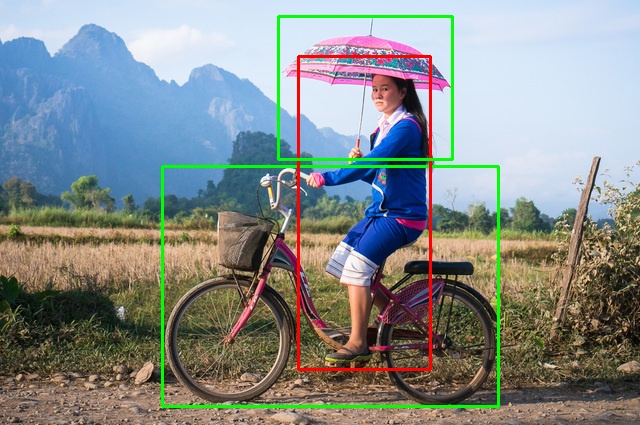}
             \centerline{\textcolor{red}{person} \textcolor{blue}{ride} \textcolor{green}{bicycle} }
             \centerline{\textcolor{red}{person} \textcolor{blue}{hold} \textcolor{green}{umbrella}} 
		\end{minipage}
	}\vspace{-10pt}
	\subfigure{
		\begin{minipage}[t]{0.2\textwidth}
			\centering
            \includegraphics[width=0.9\linewidth]{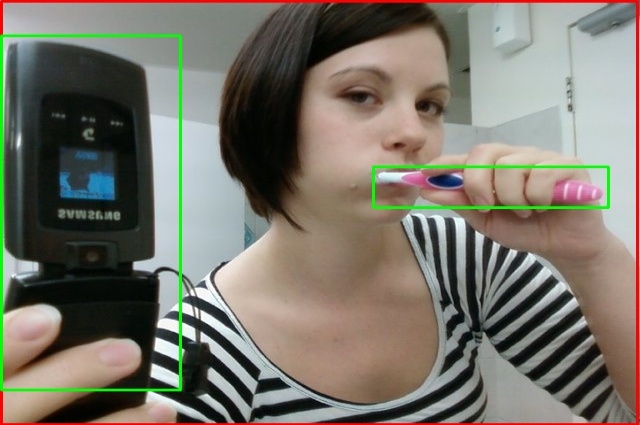}
            \centerline{\textcolor{red}{person} \textcolor{blue}{hold} \textcolor{green}{phone}} 
             \centerline{\textcolor{red}{person} \textcolor{blue}{brush} \textcolor{green}{toothbrush} }
		\end{minipage}
	}
    \subfigure{
		\begin{minipage}[t]{0.2\textwidth}
			\centering
            \includegraphics[width=0.9\linewidth]{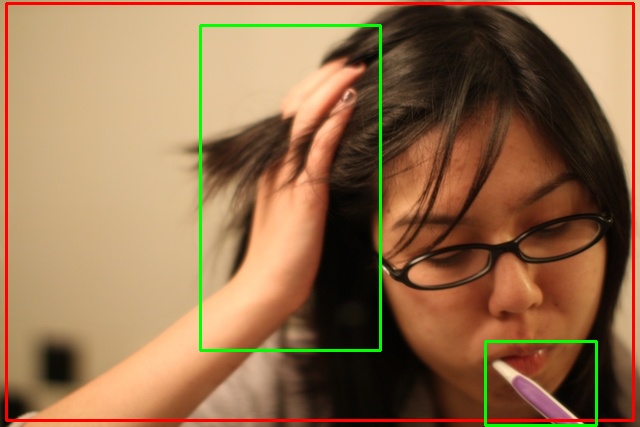}
            \centerline{\textcolor{red}{person} \textcolor{blue}{scratch} \textcolor{green}{hair}} 
             \centerline{\textcolor{red}{person} \textcolor{blue}{brush} \textcolor{green}{toothbrush} }
		\end{minipage}
	}
    \subfigure{
		\begin{minipage}[t]{0.2\textwidth}
			\centering
            \includegraphics[width=0.9\linewidth]{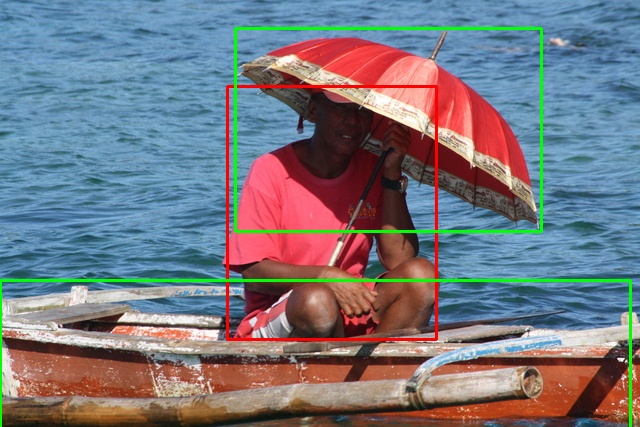}
             \centerline{\textcolor{red}{person} \textcolor{blue}{sit} \textcolor{green}{boat} }
             \centerline{\textcolor{red}{person} \textcolor{blue}{hold} \textcolor{green}{umbrella}} 
		\end{minipage}
	}
 
	\subfigure{
        \rotatebox{90}{\scriptsize{Many VS One}}
		\begin{minipage}[t]{0.2\linewidth}
			\centering
			\includegraphics[width=0.9\linewidth]{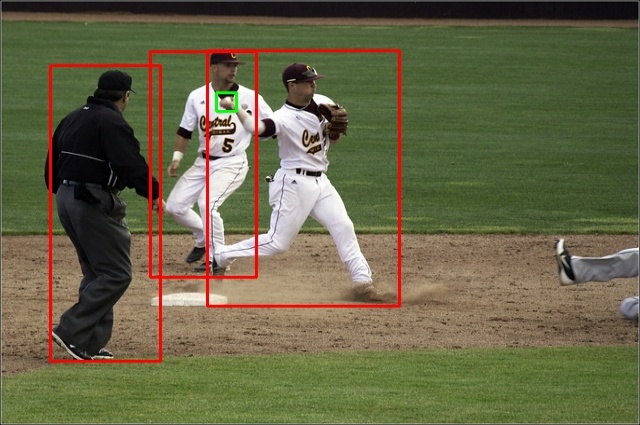}
            \centerline{\textcolor{red}{person(1)} \textcolor{blue}{hold} \textcolor{green}{baseball}} 
             \centerline{\textcolor{red}{person(1)} \textcolor{blue}{look} \textcolor{green}{person(2)} }
		\end{minipage}
	}\vspace{-10pt}
	\subfigure{
		\begin{minipage}[t]{0.2\textwidth}
			\centering
			\includegraphics[width=0.9\linewidth]{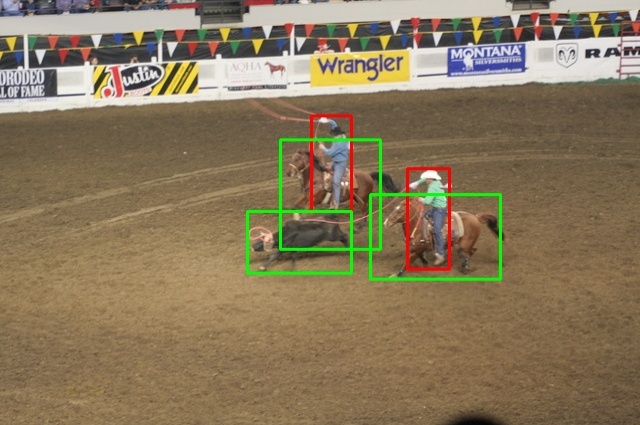}
            \centerline{\textcolor{red}{person} \textcolor{blue}{ride} \textcolor{green}{horse}} 
             \centerline{\textcolor{red}{person} \textcolor{blue}{lasso} \textcolor{green}{cow} }
		\end{minipage}
	}
    \subfigure{
		\begin{minipage}[t]{0.2\textwidth}
			\centering
			\includegraphics[width=0.9\linewidth]{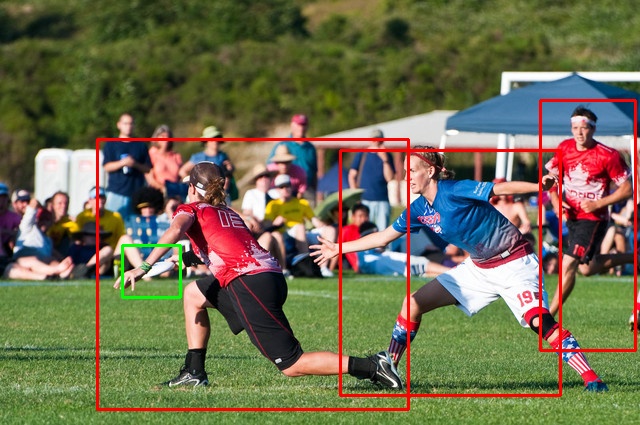}
            \centerline{\textcolor{red}{person} \textcolor{blue}{catch} \textcolor{green}{kite}} 
             \centerline{\textcolor{red}{person} \textcolor{blue}{fly} \textcolor{green}{kite} }
		\end{minipage}
	}
    \subfigure{
		\begin{minipage}[t]{0.2\textwidth}
			\centering
			\includegraphics[width=0.9\linewidth]{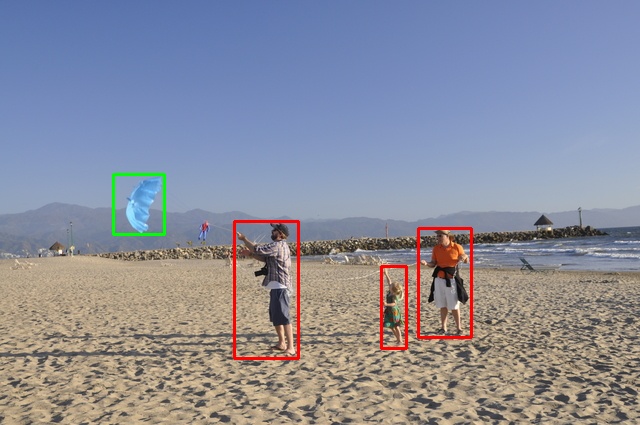}
            \centerline{\textcolor{red}{person} \textcolor{blue}{ride} \textcolor{green}{horse}} 
             \centerline{\textcolor{red}{person} \textcolor{blue}{lasso} \textcolor{green}{cow} }
		\end{minipage}
	}
 
    \subfigure{
        \rotatebox{90}{\scriptsize{Many VS Many}}
		\begin{minipage}[t]{0.2\textwidth}
		\centering
	   \includegraphics[width=0.9\linewidth]{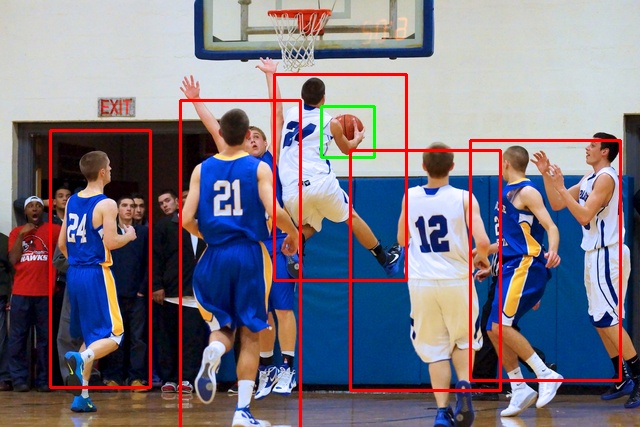}
             \centerline{\textcolor{red}{person} \textcolor{blue}{look} \textcolor{green}{baseball}}
		 \centerline{\textcolor{red}{person(1)} \textcolor{blue}{hold} \textcolor{green}{baseball} }
   \end{minipage}
	}
	\subfigure{
		\begin{minipage}[t]{0.2\textwidth}
		\centering
		\includegraphics[width=0.9\linewidth]{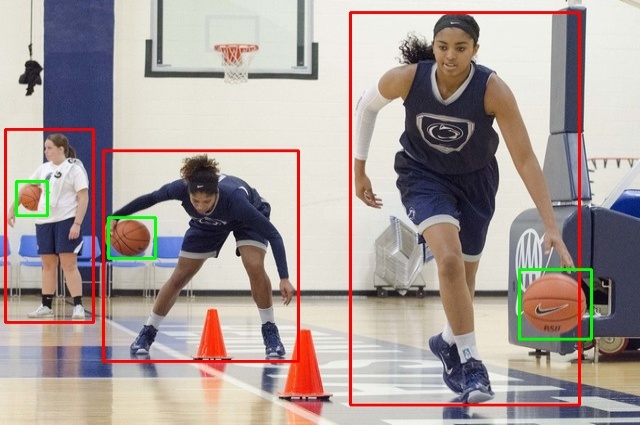}
        \centerline{\textcolor{red}{person} \textcolor{blue}{hold} \textcolor{green}{baseball} }
		\centerline{\textcolor{red}{person} \textcolor{blue}{dribble} \textcolor{green}{baseball} }
		\end{minipage}
	}
    \subfigure{
		\begin{minipage}[t]{0.2\textwidth}
		\centering
		\includegraphics[width=0.9\linewidth]{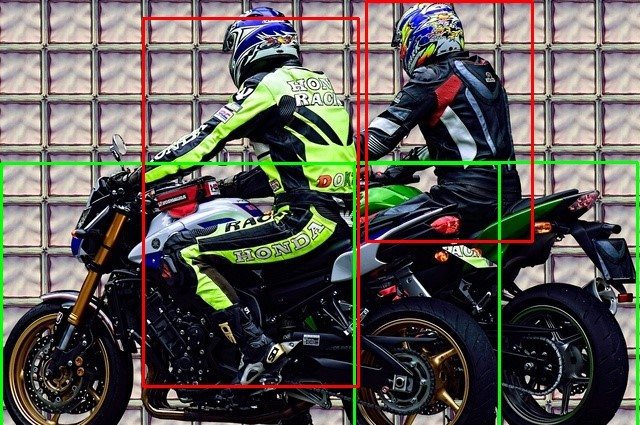}
        \centerline{\textcolor{red}{person(1)} \textcolor{blue}{ride } \textcolor{green}{motorcycle} }
		\centerline{\textcolor{red}{person(2)} \textcolor{blue}{ride} \textcolor{green}{motorcycle} }
		\end{minipage}
	}
    \subfigure{
		\begin{minipage}[t]{0.2\textwidth}
		\centering
		\includegraphics[width=0.9\linewidth]{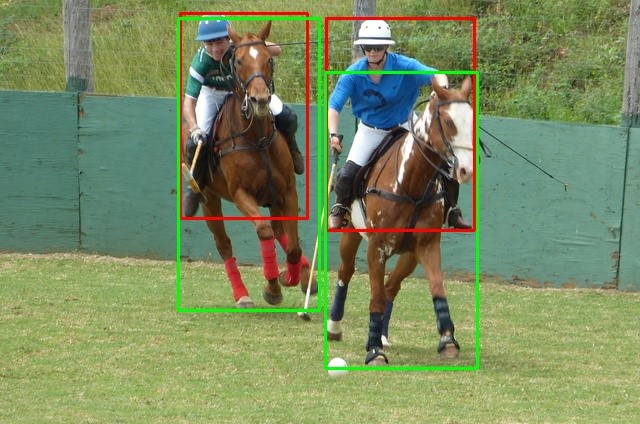}
        \centerline{\textcolor{red}{person(1)} \textcolor{blue}{ride} \textcolor{green}{horse} }
		\centerline{\textcolor{red}{person(2)} \textcolor{blue}{ride} \textcolor{green}{horse} }
		\end{minipage}
	}
 
	% 添加题注，即对这个图片的说明
	\caption{Visualization of different interaction relationships. \textbf{One VS One}: person interact with a single objects; \textbf{One VS Many}: person interact with multiple objects; \textbf{Many VS One}: person interact with a single object; \textbf{Many VS Many}: person interact with multiple objects.}
	\label{fig:result_include1}
\end{figure*}
\textbf{Datasets.} We evaluate our model on two widely-uesd benchmarks, HICO-DET~\cite{chao2018learning} and V-COCO~\cite{gupta2015visual}. 
HICO-Det contains 47,776 images, of which 38,118 and 9658 images are used for training. It includes 600 HOI triplets, where 138 triplets are rare categories less than 10 training instances, and the remaining 462 categories are non-rare. HICO-Det also provides a zero-shot detection setting by holding out 120 rare interactions. V-COCO is a subset of the COCO dataset and consists of 10,396 images, with 5,400 for training and 4,964 for testing. It has 29 action categories, including 4 annotations without any interaction with objects. 

\textbf{Zero-shot Data Setups.} We conduct experiments on the HICO-Det for zero-shot HOI detection, mainly using the following approaches: Rare First Unseen Combination (RF-UC), Non-rare First Unseen Combination (NF-UC), Unseen Verb (UV), Unseen Object (UO) and Unseen Combination(UC). In the UC setting, the training data includes all categories of objects and verbs but lacks some HOI triplet categories. We assess 120 unseen categories and 480 seen categories out of a total of 600 categories. The RF-UC selects the tail HOIs as unseen categories, whereas the NF-UC prefers head categories. In the UO setting, we select 12 unseen objects out of a total of 80 objects to define unseen HOIs. Additionally, we introduce a UV setting, where 20 verbs from a total of 117 verbs are randomly selected to construct 84 unseen and 516 seen HOIs.

\textbf{Evaluation Metric.} We evaluate our model using the mean Average Precision (mAP) as metric, a prediction is considered as true positive if the predicted human and object bounding boxes have an IoU of at least 0.5 with the ground truth, and the predicted interaction category matches the correct category. We report the standard mAP for HOI detection, dividing interactions into non-rare, rare, and unseen cases based on their occurrences in the training set. The mAP measures HOI detection performance with a threshold of 0.5 for the IoU between predicted and ground-truth bounding boxes.

\textbf{Implementation Details.} We use pre-trained DETR with ResNet-50 as the backbone network. The visual encoder is based on ViT-32/B CLIP, and during training, the parameters of CLIP remain unchanged. Our model's encoder and decoder have 3 layers with 64 queries, except the verb extraction decoder has 1 layer. We train the model using the AdamW optimizer with a specified interval for adjusting the learning rate. The training epochs is set to 90, with a gradual decrease in the learning rate after the 60th epoch. All experiments are conducted with a batch size of 8 on 4 NVIDIA A6000 GPUs.

\subsection{Effectiveness for HOI Detection}
\textbf{Zero-shot Detection.} We conduct various zero-shot setting experiments on the HICO-DET dataset, and the results are shown in Table~\ref{tab:Zero-Shot}. Our model outperforms existing state-of-the-art methods, demonstrating strong performance competitiveness. Specifically, under the RF-UC and NF-UC settings, our model's relative mean average precision (mAP) for unseen categories exceeds that of EoID~\cite{wu2023end} by 23.26$\%$ and 7.91$\%$, respectively. In the UA setting, our model outperforms the latest work, HOICLIP~\cite{10204103}, with a 1.03 mAP improvement for unseen types. In both full and unseen categories, our model achieves 1.07 and 2.5 mAP improvements relative to EOID~\cite{wu2023end}. In the case of unseen objects, our model's performance surpasses that of GEN-VLKT~\cite{9878997} by 3.21 mAP.

\textbf{Fully Supervised Detection.} To verify the generalization ability of the model, we conducted fully supervised experiments on HICO-DET and V-COCO. As tabulated in Table~\ref{tab:hico}, our model achieves remarkable performance, exceeding GEN-VLKT and HOICLIP~\cite{10204103} by 3.01 mAP and 1.14 mAP for full categories, and by 1 mAP and 0.36 mAP for rare categories. This indicates that our model can handle the long-tailed distribution of HOI well. For the V-COCO, we achieve 63.9 role AP on Scenario 1 and 65.0 role AP on Scenario 2 surpassing previous methods.

\textbf{Robustness to Distributed Data.} We investigate the robustness of the proposal under different data quantities. We decrease the proportion of training data from 100$\%$ to 15$\%$, and we seek to achieve less performance loss in HOI detection. Compared to the state-of-the-art GEN-VLKT in Table~\ref{tab:data}, the proposed method achieves competitive performance in detecting both non-rare and rare categories. Furthermore, Table~\ref{tab:data} highlights the improvements achieved by our model at various volumes of data. At 25$\%$ training data, our model exhibits a 78.41$\%$ increase in mAP gain for rare HICO-DET.
\subsection{Ablation Studies}
\textbf{Network Architecture Analysis.} We conduct ablation experiments for each module on the HICO-DET dataset under the UV setting and the results are shown in Table~\ref{tab:Architecture}. The GEN-VLKT without the knowledge distillation component.is regarded as the baseline. First, we examine the effect of CLIP and the result shows a 13.8 mAP improvement in HOI detection for Unseen categories. Thus, the visual and linguistic knowledge extracted by CLIP enable to learn deeper interaction understanding. Next, we replace the encoder with the proposed Ho-Pair Encoder for using fine-grained visual features. As a result, we observe that the results come up to 30.99 mAP and 32.02 mAP in the Full and Seen categories, respectively. Finally, we add a verb feature learning to capture verb-related features, and the performance is further advanced to 31.85 mAP in full categories, which demonstrates the necessity of verb feature learning in the zero-shot HOI task.

\textbf{Reconstruction Loss Setting.} As shown in Table~\ref{tab:LOSS}, we compare two loss types, \textit{i.e.}, $L_1$ loss and $L_2$ loss, as the reconstruction loss. It demonstrates that the reconstruction loss is indispensable for making knowledge transfer effective. If exclusively employing $L_1$ loss, our model has demonstrated superior performance and achieved 1.05 mAP gain, surpassing the performance exhibited by only employing the $L_2$ loss. When combining $L_1$ and $L_2$ losses through average summation, the performance is still inferior compared to only applying $L_1$ loss.

\textbf{Verb Extraction Decoder Layer Selection.} We examine the effect of verb extraction decoder layer number on the verb features update and conduct experiments in the NF-UC setting. In Table~\ref{tab:layers}, we find that a single layer exhibits superior performance. While increasing the number of layers, there is no consistent improvement.

\subsection{Qualitative Visualization Results}
As shown in Figure~\ref{Qualitative}, we illustrate the characteristics of our model by visualizing the attention feature maps of the decoder. Our framework can effectively infer human-object interaction relationships in unseen interaction categories. We observe the attention maps from the Interaction Representation Decoder focus on a broader range of interaction-related regions, while the Verb Extraction Decoder mainly emphasizes the interaction target area. Additionally, taking the detection of ``wash-bus``, as an example, it indicates that our model is capable of perceiving interactions between indirectly connected humans and objects. In Figure~\ref{fig:result_include1}, we note that our model has the capability to detect various types of interactions.

\section{CONCLUSION}
The present paper introduces a novel one-stage framework, known as KI2HOI, for zero-shot HOI detection. KI2HOI effectively integrates visual-linguistic knowledge from CLIP to augment the interaction representation and recognition unknown instances of HOIs. We also introduce Ho-Pair Encoder to generate contextual spatial features via additive attention mechanism and extract verb features from associated queries. To enhance the understanding of interactions, we adopt a novel interaction representation decoder via text embeddings of CLIP. Extensive experimental results demonstrate the effectiveness of the proposed KI2HOI framework.

\section{CONFLICT OF INTEREST STATEMENT}
We declare no competing financial interests or personal
relationships that could have influenced the work presented
in this paper.

\bibliographystyle{IEEEtran}
\bibliography{K2HOI}

\begin{IEEEbiography}[{\includegraphics[width=1in,height=1.25in,clip,keepaspectratio]{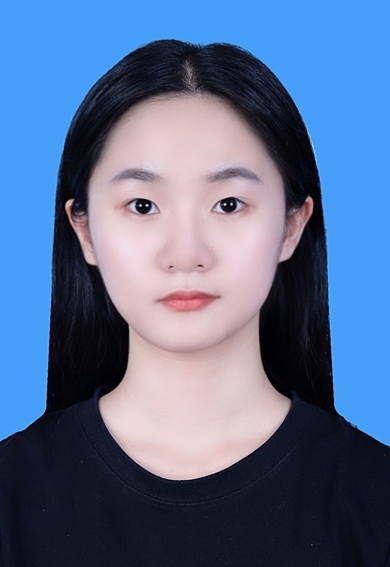}}]{Weiying Xue}
is currently pursuing an M.D degree at the School of Future Technology, South China University of Technology (SCUT), China. Her current research interests include human-object interaction, including human-object interaction detection, visual relation detection, object detection, etc.
\end{IEEEbiography}

\vspace{-5\baselineskip}

\begin{IEEEbiography}[{\includegraphics[width=1in,height=1.25in,clip,keepaspectratio]{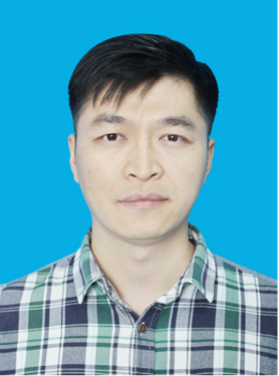}}]{Qi Liu}
is currently a Professor with the School of Future Technology at South China University of Technology. Dr. Liu received the Ph.D degree in Electrical Engineering from City University of Hong Kong, Hong Kong, China, in 2019. During 2018 - 2019, he was a Visiting Scholar at University of California Davis, CA, USA. From 2019 to 2022, he worked as a Research Fellow in the Department of Electrical and Computer Engineering, National University of Singapore, Singapore. His research interests include human-object interaction, AIGC, 3D scene reconstruction, and affective computing, etc. Dr. Liu has been an Associate Editor of the IEEE Systems Journal (2022-), and Digital Signal Processing (2022-). He was also Guest Editor for the IEEE Internet of Things Journal, IET Signal Processing, etc. He was the recipient of the Best Paper Award of IEEE ICSIDP in 2019.
\end{IEEEbiography}
\vspace{-5\baselineskip}

\begin{IEEEbiography}[{\includegraphics[width=1in,height=1.25in,clip,keepaspectratio]{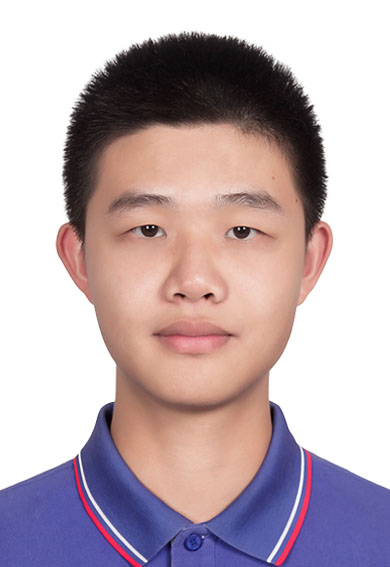}}]{Qiwei Xiong}
received the M.E. degree from the University of Technology Sydney, Sydney, Australia, in 2021. He is currently pursuing the Ph.D.degree with electronic information from South China University of Technology, Guangzhou, China. His main research interests include Human-computer interaction, loT sensing and mobile computing.
\end{IEEEbiography}

\vspace{-5\baselineskip}

\begin{IEEEbiography}[{\includegraphics[width=1in,height=1.25in,clip,keepaspectratio]{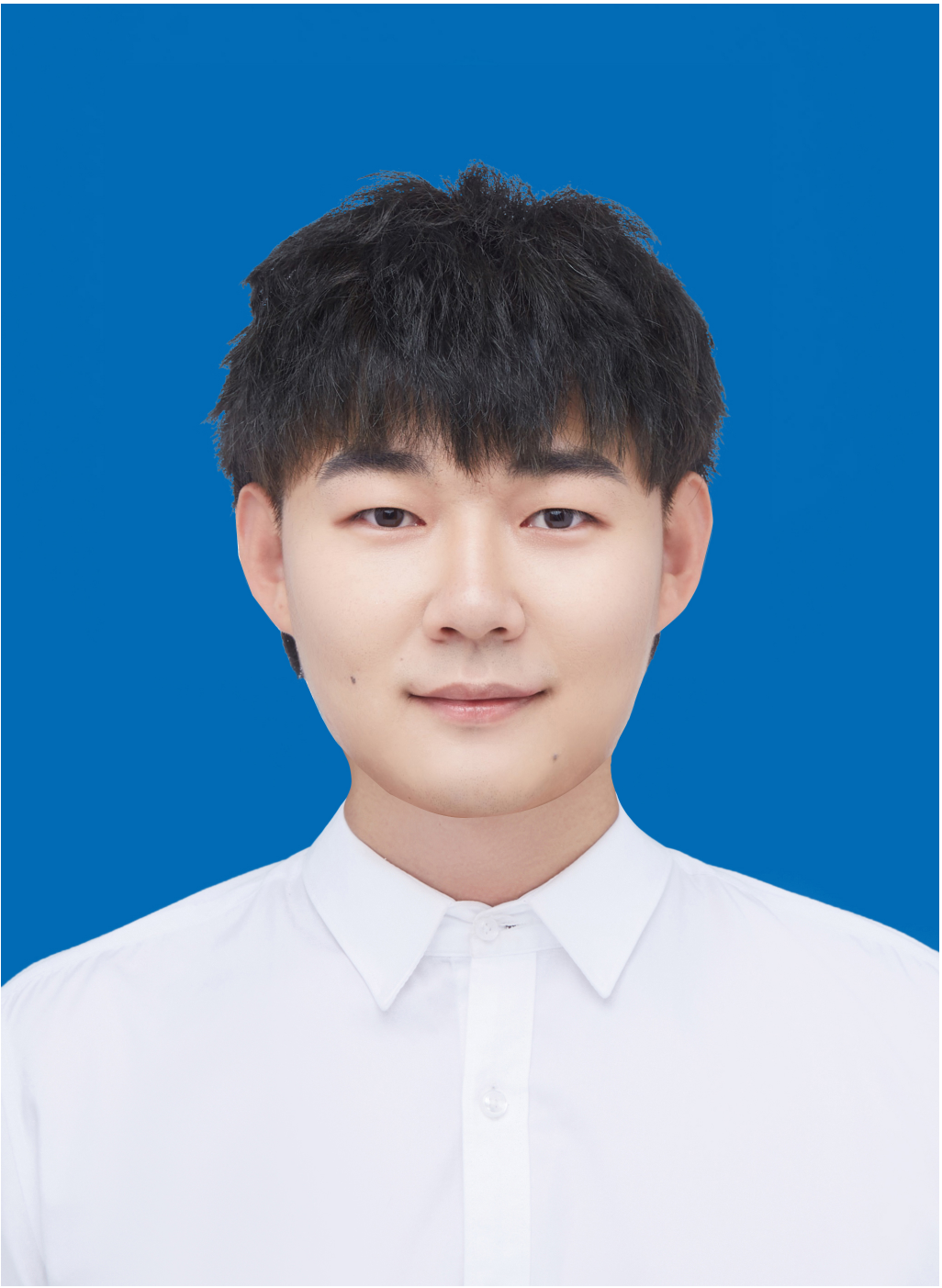}}]{Yuxiao Wang}
is currently pursuing a Ph.D degree at the School of Future Technology, South China University of Technology (SCUT), China. His research focuses on human-object interaction, including human-object interaction detection, human-object contact detection, semantic segmentation, crowd counting, etc. In addition, he has conducted research on weakly supervised learning methods.
\end{IEEEbiography}

\vspace{-5\baselineskip}

\begin{IEEEbiography}[{\includegraphics[width=1in,height=1.25in,clip,keepaspectratio]{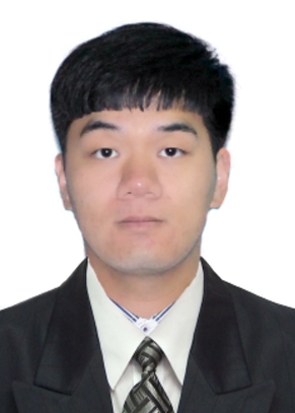}}]{Zhenao Wei}
received the D.Eng. degree from the Graduate School of Information Science and Engineering, Ritsumeikan University, Japan. He is currently a postdoctoral fellow at South China University of Technology. His research interests include game AIs, human-object interaction, and human-object contact.
\end{IEEEbiography}
\vspace{-5\baselineskip}

\begin{IEEEbiography}[{\includegraphics[width=1in,height=1.25in,clip,keepaspectratio]{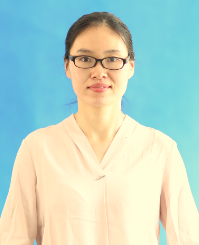}}]{Xiaofen Xing}
is currently an Associate Professor with the School of Future Technology at the South China University of Technology. Dr. Xing participated in the key project of the National Natural Science Foundation Joint Fund (Brain-like Affective Computing(2019-2022). In recent years, she has focused on deep learning-based cross-media emotion computing, covering images, expressions, speech, natural language understanding, and 3D virtual avatar interaction. It has undertaken 1 youth project of the National Natural Science Foundation and 3 science and technology plan projects such as the special research and development project of artificial Intelligence in Guangdong Province. She has published more than 20 articles in well-known journal conferences in the field of artificial intelligence, won the best paper nomination award of the well-known image processing conference ICIP, and was shortlisted for the authoritative Emotion competition paper report of ACM MM twice.
\end{IEEEbiography} 

\vspace{-3\baselineskip}

\begin{IEEEbiography}[{\includegraphics[width=1in,height=1.25in,clip,keepaspectratio]{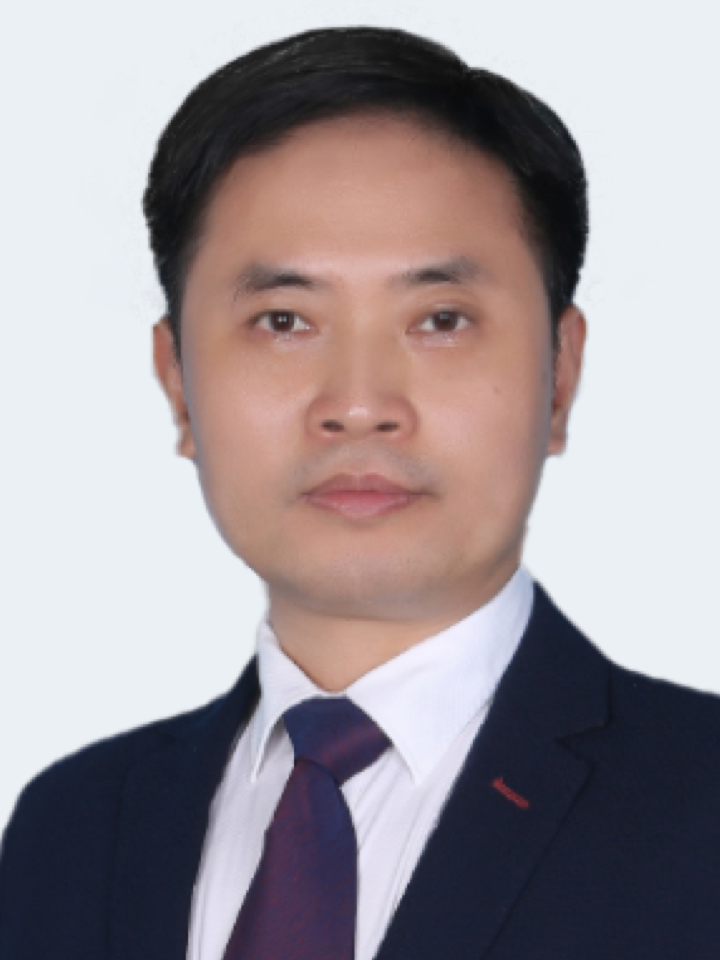}}]{Xiangmin Xu}
is the Second-Grade Professor of South China University of Technology, Doctoral Supervisor, Vice President of South China University of Technology, Dean of the school of Future Technology, Member of Electronic Information Teaching Steering Committee of Ministry of Education, Director of Engineering Research Center of Ministry of Education on Human Body Data Perception. His recent research focuses on computer vision, natural interaction, affective computing and so on, and the doctoral student he trained won the Excellent Doctoral Dissertation of the Chinese Society of Electronics in 2020. He has undertaken more than 20 projects, published more than 60 articles in top journals and conferences, including 1 ESI hot paper with high citation, 50 invention patents were authorized, and the research achievements of Computational Psychophysiology were granted with the 2nd class of National Technology Invention Award in 2018. He has published 7 papers in important educational journals and been awarded the 2nd class of the National Teaching Achievements Award for three consecutive years.
\end{IEEEbiography} 

% \newpage
% \section{Biography Section}
% If you have an EPS/PDF photo (graphicx package needed), extra braces are
%  needed around the contents of the optional argument to biography to prevent
%  the LaTeX parser from getting confused when it sees the complicated
%  $\backslash${\tt{includegraphics}} command within an optional argument. (You can create
%  your own custom macro containing the $\backslash${\tt{includegraphics}} command to make things
%  simpler here.)
 
% \vspace{11pt}

% \bf{If you include a photo:}\vspace{-33pt}
% \begin{IEEEbiography}[{\includegraphics[width=1in,height=1.25in,clip,keepaspectratio]{fig1}}]{Michael Shell}
% Use $\backslash${\tt{begin\{IEEEbiography\}}} and then for the 1st argument use $\backslash${\tt{includegraphics}} to declare and link the author photo.
% Use the author name as the 3rd argument followed by the biography text.
% \end{IEEEbiography}

% \vspace{11pt}

% \bf{If you will not include a photo:}\vspace{-33pt}
% \begin{IEEEbiographynophoto}{John Doe}
% Use $\backslash${\tt{begin\{IEEEbiographynophoto\}}} and the author name as the argument followed by the biography text.
% \end{IEEEbiographynophoto}

% \vfill

\end{document}